\pdfoutput=1
\documentclass{article}

\usepackage[margin=1in]{geometry}

\usepackage{amsmath,amssymb,amsfonts,amsthm,mathtools}
\usepackage{booktabs}
\usepackage{microtype}
\usepackage{nicefrac}
\usepackage[numbers]{natbib}
\usepackage{url}
\usepackage[hidelinks]{hyperref}
\usepackage{enumitem}
\usepackage{xcolor}
\usepackage{graphicx}

\title{Geometric Stability and Sample Complexity for Compositional Function Priors\thanks{An extended abstract of this work was accepted for presentation at the NeurIPS 2026 Workshop on Symmetry and Geometry in Neural Representations (NeurReps), a non-archival venue.}}

\author{Chris Hammond\\
  Susquehanna International Group\\
  \texttt{chris.hammond@sig.com}}

\date{November 2026}

\newtheorem{theorem}{Theorem}
\newtheorem{proposition}{Proposition}
\newtheorem{lemma}{Lemma}
\newtheorem{corollary}{Corollary}
\newtheorem{definition}{Definition}
\newtheorem{assumption}{Assumption}
\newtheorem{remark}{Remark}

\newcommand{\R}{\mathbb{R}}
\newcommand{\E}{\mathbb{E}}
\newcommand{\F}{\mathcal{F}}
\newcommand{\G}{\mathcal{G}}
\newcommand{\Hcal}{\mathcal{H}}
\newcommand{\X}{\mathcal{X}}
\newcommand{\norm}[1]{\left\lVert #1 \right\rVert}
\newcommand{\ip}[2]{\left\langle #1,#2 \right\rangle}

\newcommand{\eps}{\varepsilon}

\begin{document}
\maketitle

\begin{abstract}
We study when compositional structure yields dimension reduction in supervised learning.  We frame learning as empirical risk minimization with a function-space prior: a regularization functional selects a subset of functions with favorable approximation and metric entropy.  The models are compositions
\(
    f(x)=g(Ax+\Delta h(x)),
\)
where \(A\in\R^{k\times d}\) is a linear base map, \(\Delta h\) is a nonlinear correction, and \(g\) is an outer function; that is, \(f\) is an arbitrary smooth function of a \(k\)-dimensional representation of the input that is linear up to a correction.  We prove covering-number bounds showing that the statistical complexity of this class interpolates between an intrinsic regime---parametric in the \(kd\) entries of the linear map and nonparametric only in the representation dimension \(k\)---and an ambient \(d\)-dimensional nonparametric rate; the interpolation is controlled by the correction size \(\eps=\norm{\Delta h}\).  We then connect \(\eps\) to data geometry.  If the data lie in a small tube around a \(k\)-dimensional manifold and the regression target depends on the input only through its nearest point on that manifold, so that the target is really a function on the manifold, then locally the target factorizes exactly as \(g(Ax+\Delta h(x))\) with \(A\) a tangent chart and a correction whose Sobolev norm scales linearly with curvature at the localization scale.  Patching such local models over an atlas turns the local statement into a global one: we obtain a finite-sample excess-risk bound for a least-squares estimator over a finite net of local models, whose ambient term vanishes with curvature.  For a one-dimensional model problem we also prove a converse: on curved data, every affine compositional predictor faces an error floor unless its correction budget scales with curvature, while the matching curved model class attains the intrinsic rate.  The correction size \(\eps\) is thus the effective-dimension parameter linking curvature to statistical complexity.

\end{abstract}

\section{Introduction}

Compositional structure is central to modern machine learning. Deep networks implement layered compositions \citep{mallat2016understanding,poggio2017why}, multi-index models posit functions of low-dimensional projections \citep{li1991sliced,xia2002adaptive}, and representation learning seeks feature maps that reduce effective dimension \citep{belkin2003laplacian}. A fundamental question is therefore:

\begin{quote}
\emph{When does compositional structure actually reduce statistical complexity from an ambient dimension $d$ to an intrinsic dimension $k \ll d$?}
\end{quote}

We approach this question through a function-prior perspective. Given data $\{(x_i, y_i)\}_{i=1}^n$, many learning procedures can be idealized as solving
\begin{equation}
\label{eq:erm-prior}
\hat f \in \arg\min_{f \in \mathcal{F}} \frac{1}{n} \sum_{i=1}^n L(f(x_i), y_i) + \lambda \Phi(f)
\end{equation}
where $\Phi$ encodes prior assumptions on the unknown regression function. In both Bayesian and frequentist settings, two quantities govern generalization: (i) \emph{approximation}, meaning whether the prior places sufficient mass or capacity near the truth, and (ii) \emph{complexity}, often measured by metric entropy or Rademacher complexity.

We study these questions for compositional models of the form
\begin{equation}
\label{eq:model}
f(x) = g(h(x)), \quad h(x) = Ax + \Delta h(x)
\end{equation}
where $A \in \mathbb{R}^{k \times d}$ is a linear base map, $\Delta h : \mathbb{R}^d \to \mathbb{R}^k$ is a nonlinear correction, and $g : \mathbb{R}^k \to \mathbb{R}$ is an outer function. The quantity
\[
\varepsilon = \|\Delta h\|
\]
(a Sobolev norm, made precise in Definition~\ref{def:class}) measures the deviation from a purely linear multi-index model. We interpret $\varepsilon$ as the \emph{effective nonlinearity} of the representation, and show that it governs both statistical complexity and its dependence on geometry.

Our main conceptual contribution is to identify $\varepsilon$ as the quantity mediating between compositional structure and statistical behavior. We show that the complexity of the class interpolates between intrinsic and ambient regimes as $\varepsilon$ varies: when $\varepsilon$ is small, the class behaves like a $k$-dimensional nonparametric model, up to a parametric term in the $kd$ entries of the linear map, while for large $\varepsilon$, ambient $d$-dimensional effects dominate. Crucially, the ambient contribution enters only through $\varepsilon$, isolating the mechanism by which compositional structure can reduce effective dimension.

We then provide a geometric explanation for when $\varepsilon$ is small. If the data lie in a small tube around a smooth $k$-dimensional manifold with bounded curvature $\kappa$ and controlled higher-order regularity, then the feature map inverting the nearest-point projection onto the manifold differs from a tangent linear map by a correction of size proportional to $\kappa$, and the regression target factorizes exactly through this feature map. Together these results show that curvature at a given scale bounds $\varepsilon$ linearly (Proposition \ref{prop:geometry}), $\varepsilon$ alone controls the ambient part of the metric entropy (Theorem \ref{thm:covering}), and the entropy controls the excess risk (Theorem \ref{thm:risk}); composing them gives rates whose ambient term vanishes with curvature (Corollary \ref{cor:rates}).

\paragraph{Why this matters.}
Many modern models rely on compositional structure to achieve strong performance in high dimensions, but theoretical explanations often assume low intrinsic dimension a priori. Our results instead identify a concrete mechanism—small correction size $\eps$—under which such behavior emerges from geometric regularity, providing a bridge between representation learning and classical nonparametric theory.

\paragraph{Contributions.}
Our contributions are as follows:
\begin{itemize}
\item We formalize compositional learning in a function-prior framework, distinguishing explicit regularization from the implicit priors induced by architecture and optimization.
\item We prove covering-number bounds for compositional classes that explicitly separate intrinsic and ambient contributions, with the ambient term controlled by the correction size $\varepsilon$.
\item We show that, for data in a tube around a manifold with bounded curvature and controlled higher-order regularity, the target factorizes exactly through a nearly linear feature map with $\varepsilon$ proportional to curvature at the localization scale.  This supplies both the approximation and the complexity halves of the argument, providing a geometric mechanism for dimension reduction.
\item We combine the entropy and geometric results into a finite-sample excess-risk bound for a least-squares estimator over a finite net of local compositional models, exhibiting parametric, intrinsic-dimensional, and curvature-driven ambient contributions, and we analyze the choice of the localization scale.
\item We prove a class-level floor and separation theorem for a one-dimensional model problem: on data in a tube around a circle, every affine compositional predictor incurs an $L^2$ error floor unless its correction budget scales with curvature, while the matching circle-model class attains the intrinsic rate.  The correction budget is thus necessary, not merely sufficient.  In any dimension we prove a representation floor: no affine feature map with budget below the curvature scale serves all targets on a curved chart, and the target that defeats a given feature map is explicit.  A fixed scalar target can evade the floor when $k\ge2$, and we explain why.
\end{itemize}

\paragraph{Relation to prior work.}
Classical multi-index and manifold models assume low intrinsic dimension a priori. In contrast, our analysis identifies a quantitative condition—small correction size $\varepsilon$—under which a compositional class \emph{behaves} as low-dimensional. Our work is also related to geometric stability results \citep{bietti2021deep}, but differs in that we analyze explicit compositional function classes and isolate the role of nonlinear correction rather than invariance alone.

Together, these results provide a quantitative account of when compositional function priors behave as low-dimensional models, and identify curvature as a key driver of this behavior.
\section{Function priors and compositional regularization}

The prior viewpoint behind \eqref{eq:erm-prior} is useful because it separates two questions.  The first is approximation: does the prior or regularizer make the true function inexpensive?  The second is complexity: how large is the set of functions with comparable cost?  For a sublevel set
\[
    \F_B=\{f\in\F:\Phi(f)\le B\},
\]
generalization can be controlled through metric entropy, e.g., \(\log N(\delta,\F_B,L^2)\), or through Rademacher complexity \citep{dudley1967sizes,bartlett2002rademacher,wainwright2019high}.  In Bayesian nonparametrics, analogous quantities govern posterior contraction through prior mass near the truth and entropy of sieves \citep{ghosal2000convergence,vandervaart2008rates}.

For compositional models, a natural explicit prior is obtained by assigning separate costs to the components.  If \(f=g\circ h\) and independent priors are placed on \(g\) and \(h\), then the negative log-prior is additive at the level of components.  For the model \eqref{eq:model}, this motivates penalties of the form
\begin{equation}
    \norm{A}_F^2+\norm{\Delta h}_{\Hcal_h}^2+\norm{g}_{\Hcal_g}^2.
    \label{eq:additive-prior}
\end{equation}
The additivity in \eqref{eq:additive-prior} should not be confused with additivity of the induced function class.  Composition couples the components nonlinearly, and the statistical complexity of the resulting class is not merely the sum of the complexities of the pieces.  This coupling is the main source of both the usefulness and the difficulty of compositional priors.

Appendix \ref{app:penalty} records how this penalty relates to RKHS regularization and the finite-dimensional, nonconvex form of the resulting optimization problem.

\section{Compositional classes and entropy interpolation}

We now state the main statistical bound.  The theorem is formulated for Sobolev-type component classes because this makes the dimension dependence explicit.  Other entropy estimates can be substituted when different component priors are used \citep{cucker2002mathematical,zhou2002covering,steinwart2020closer}.

The key point is that the ambient-dimensional contribution arises only through perturbations of the linear representation, and is therefore controlled by the correction radius $\eps$.

\begin{assumption}[Domain and boundedness]
\label{assump:domain}
The inputs satisfy \(x\in\X\subset\R^d\), where \(\X\) is bounded and \(\sup_{x\in\X}\norm{x}\le R\).  The empirical and population \(L^2\) norms are computed with respect to the sample and the data distribution, respectively.  Functions are uniformly bounded on the relevant domains.
\end{assumption}

\begin{assumption}[Smooth component classes]
\label{assump:smooth}
The correction \(\Delta h\) belongs to a vector-valued Sobolev ball of smoothness \(s_h>d/2\), radius \(\eps\), on \(\R^d\).  The outer function \(g\) belongs to a Sobolev ball of smoothness \(s_g>k/2+1\), radius \(B_g\), on a bounded subset of \(\R^k\).  In particular, \(g\) is Lipschitz with Lipschitz constant at most \(L_g\lesssim B_g\).
\end{assumption}

\begin{definition}[Linear-base compositional class]
\label{def:class}
Given \(B_A,B_g,\eps>0\), define
\begin{equation}
\F_{B_A,B_g,\eps}
=\left\{g(Ax+\Delta h(x)):
\norm{A}_F\le B_A,\ \norm{\Delta h}_{W^{s_h,2}}\le \eps,
\ \norm{g}_{W^{s_g,2}}\le B_g\right\}.
\label{eq:function-class}
\end{equation}
\end{definition}

For a class $\G$ and a norm $\rho$, write $N(\delta,\G,\rho)$ for the $\delta$-covering number of $\G$, the smallest number of $\rho$-balls of radius $\delta$ whose union contains $\G$; its logarithm $\log N(\delta,\G,\rho)$ is the metric entropy.  Two norms are used: the empirical norm $L_n^2$, $\rho(f,f')^2=\frac1n\sum_{i=1}^n(f(x_i)-f'(x_i))^2$, and the sup norm $\norm{\cdot}_\infty$.

\begin{theorem}[Covering interpolation for compositional priors]
\label{thm:covering}
Under Assumptions \ref{assump:domain}--\ref{assump:smooth}, up to constants depending on \(d,k,s_h,s_g,R\) and boundedness parameters, the empirical covering number of \(\F_{B_A,B_g,\eps}\) satisfies
\begin{equation}
\log N(\delta,\F_{B_A,B_g,\eps},L_n^2)
\lesssim
kd\log\!\left(\frac{C B_A B_g}{\delta}\right)
+
\left(\frac{C B_g\eps}{\delta}\right)^{d/s_h}
+
\left(\frac{C B_g}{\delta}\right)^{k/s_g}.
\label{eq:covering-bound}
\end{equation}
Consequently, standard entropy-integral bounds imply that the Rademacher complexity is controlled by the sum of a finite-dimensional linear-map term, an ambient-dimensional correction term whose radius is proportional to \(\eps\), and an intrinsic-dimensional outer-function term.
\end{theorem}

\begin{proof}[Proof outline; details in Appendix \ref{app:covering}]
Cover the matrix ball \(\{A:\norm{A}_F\le B_A\}\) in Frobenius norm, giving a logarithmic term of order \(kd\log(CB_A/\delta)\).  Cover the correction class in empirical \(L^2\), using Sobolev entropy
\(\log N(u,\eps B_{W^{s_h,2}},L_n^2)\lesssim (\eps/u)^{d/s_h}\).  Cover the outer class using
\(\log N(v,B_g B_{W^{s_g,2}},L_n^2)\lesssim (B_g/v)^{k/s_g}\).  If \(h=Ax+\Delta h\) and \(\widetilde h=\widetilde A x+\widetilde{\Delta h}(x)\), then Lipschitzness of \(g\) gives
\[
\norm{g\circ h-g\circ \widetilde h}_{L_n^2}
\le L_g\norm{h-\widetilde h}_{L_n^2}.
\]
The remaining error is the approximation of \(g\) on the covered feature points.  Choosing component accuracies so that the three errors sum to \(\delta\) yields \eqref{eq:covering-bound}.  A more detailed argument, recording where each assumption enters, is given in Appendix \ref{app:covering}.
\end{proof}

\begin{remark}[Sample-complexity regimes]
\label{rem:sample}
Up to logarithmic factors and constants, the entropy bound yields two qualitative regimes.  If \(\eps\) is small enough that the correction term in \eqref{eq:covering-bound} is negligible at the target accuracy, the effective statistical complexity is governed by \(kd\) and the intrinsic \(k\)-dimensional entropy of \(g\).  If \(\eps\) is not small, the ambient \(d\)-dimensional correction term can dominate.  Thus \(\eps\) controls interpolation between intrinsic and ambient behavior.
\end{remark}

This statement intentionally avoids claiming minimax optimality for the composed class.  For ordinary Sobolev or RKHS balls, entropy methods can recover classical rates under appropriate assumptions \citep{yang1999information}.  For the coupled compositional class, matching lower bounds require additional work.  The contribution here is the explicit dependence on the correction radius \(\eps\), which isolates the mechanism by which a compositional prior can behave as low-dimensional. The ambient-dimensional contribution enters only through $\eps$. In particular, as $\eps \rightarrow 0$, the bound reduces to a $k$-dimensional regime.

\begin{remark}[The ambient term is not an artifact]
\label{rem:lower}
The \(\eps\)-term in \eqref{eq:covering-bound} reflects genuine capacity rather than slack in the argument.  If \(B_g\) exceeds a fixed constant, the outer class contains a localized coordinate map \(z\mapsto z_1\), and taking \(A=0\) shows that \(\F_{B_A,B_g,\eps}\) contains a full \(d\)-dimensional Sobolev ball of radius proportional to \(\eps\), provided \(\eps\) is at most a fixed constant so that, by the sup-norm bound of Appendix \ref{app:covering}, all feature values remain in the region where the localized map coincides with the coordinate.  Classical entropy lower bounds for Sobolev balls \citep{wainwright2019high} then give \(\log N(\delta,\F_{B_A,B_g,\eps},L^2)\gtrsim (c\eps/\delta)^{d/s_h}\) in the population norm for suitable designs.  Hence for large \(\eps\) the class genuinely carries ambient \(d\)-dimensional complexity, and the interpolation of Theorem \ref{thm:covering} describes real behavior of the class, not looseness of the upper bound.  
\end{remark}

\section{Geometry controls the correction size}

The entropy bound reduces the statistical question to a geometric one: when is \(\eps\) small?  We answer this locally, for data supported in a small tube around a smooth manifold.  The off-manifold setting is the substantive one: for data supported exactly on a manifold, the tangent projection alone is already a valid chart and no correction is needed (Remark \ref{rem:necessity}).  It is noisy or off-manifold data that force a genuinely nonlinear correction, and curvature controls how large that correction must be.

Fix a base point $\bar x\in M$ and a scale $r_0>0$.  After a rigid motion, $\bar x=0$ and $T_{\bar x}M=\R^k\times\{0\}$, so that near $0$ the manifold is the graph of $\varphi(u)=(u,\psi(u))$ with $\psi(0)=0$ and $D\psi(0)=0$.  Such \emph{adapted coordinates} exist for every embedded submanifold; all geometric statements below are made in them.

\begin{assumption}[Geometric regularity at scale $r_0$]
\label{assump:geom}
$M\subset\R^d$ is a $C^{s_h+1}$ embedded $k$-dimensional submanifold with reach at least $\rho>0$ \citep{niyogi2008finding}, and in adapted coordinates at $\bar x$ the graph representation $\varphi(u)=(u,\psi(u))$ is valid for all $u\in B(0,r_0)\subset\R^k$.
\end{assumption}

\begin{definition}[Curvature aggregate at scale $r_0$]
\label{def:curvature}
Under Assumption \ref{assump:geom}, for $2\le m\le s_h+1$ set
\[
    \kappa_m \;:=\; \sup_{\norm{u}\le r_0}\norm{D^m\psi(u)},
    \qquad
    \bar\kappa \;:=\; \max_{2\le m\le s_h+1}\,\kappa_m\, r_0^{\,m-2}.
\]
\end{definition}

The parameter $\kappa_2$ bounds $\norm{D^2\psi}$ over the chart; at $\bar x$ this is the second fundamental form of $M$ in these coordinates, and over the chart it bounds the second fundamental form up to a factor $1\pm O(\bar\kappa r_0)$.  The higher-order parameters $\kappa_m$ do not enlarge $\bar\kappa$ whenever $\kappa_m\le\kappa_2\,r_0^{\,2-m}$, i.e.\ whenever higher-order oscillation is dominated at scale $r_0$.  In that regime $\bar\kappa=\kappa_2$ and the statements below read as curvature-to-correction bounds in the classical sense; in general only $\kappa_2$ is curvature in the classical sense, and the honest statement is that $\bar\kappa$, curvature at scale $r_0$, controls the correction.  Throughout the geometric statements we take $s_h$ and $s_g$ to be integers; for fractional smoothness indices, replace them by their integer ceilings.

\begin{proposition}[Curvature controls the correction, with exact factorization]
\label{prop:geometry}
Let Assumption \ref{assump:geom} hold with the smoothness index $s_h>d/2$ of Assumption \ref{assump:smooth}, fix a tube radius $0\le\tau\le\min(\rho/2,\,r_0)$, and suppose $\bar\kappa\, r_0\le c_0$, where $c_0=c_0(d,k,s_h)\in(0,1/10]$ is the constant fixed in Appendix \ref{app:geometry}: small enough that the inverse function theorem applies to the rescaled tube map, and at most $1/10$ as the proof of Corollary \ref{cor:atlas} requires.  Let $A\in\R^{k\times d}$ have orthonormal rows spanning $T_{\bar x}M$, so that $\norm{A}_F=\sqrt k$.  Let $\pi$ denote the nearest-point projection onto $M$ and define the local tube
\[
    T_{r_0,\tau}
    :=\bigl\{x\in\R^d:\ \mathrm{dist}(x,M)\le\tau,\
    \pi(x)\in\varphi\bigl(B(0,r_0/2)\bigr)\bigr\}.
\]
Then there exists $\Delta h\in W^{s_h,2}(\R^d;\R^k)$, supported in $B(0,2r_0)$, such that the feature map $h(x)=Ax+\Delta h(x)$ has the following properties.
\begin{enumerate}
\item[(i)] \emph{(Well-specified feature map.)}  For every $x\in T_{r_0,\tau}$,
\[
    h(x) \;=\; u(\pi(x)),
\]
where $u(\cdot)$ denotes the graph coordinate, $u(\varphi(u))=u$.  That is, $h$ inverts the manifold parameterization after projecting out the normal directions: it is constant on normal fibers, and its restriction to $M\cap\varphi(B(0,r_0/2))$ is a chart with bi-Lipschitz distortion $1\pm C\bar\kappa r_0$.
\item[(ii)] \emph{(Correction size.)}  For $0\le m\le s_h$,
\begin{equation}
    \norm{D^m\Delta h}_{L^2(\R^d)}
    \;\le\; C\,\bar\kappa\; r_0^{\,d/2+2-m},
    \label{eq:curvature-bound-safe}
\end{equation}
and consequently
\[
    \eps \;:=\;\norm{\Delta h}_{W^{s_h,2}(\R^d)}
    \;\le\; C(d,k,s_h)\;\bar\kappa\;
    r_0^{\,d/2}\max\bigl(r_0^{\,2-s_h},\,r_0^{\,2}\bigr).
\]
In particular the dependence on the curvature aggregate $\bar\kappa$ is linear.
\item[(iii)] \emph{(Exact factorization of the target.)}  Suppose the target satisfies $f=f_0\circ\pi$ on $T_{r_0,\tau}$ for some $f_0:M\to\R$ with $\norm{f_0\circ\varphi}_{C^{s_g}(B(0,r_0))}\le B_g$, the uniform norm on derivatives up to order $s_g$.  Then there exists a compactly supported $g$ with $\norm{g}_{W^{s_g,2}(\R^k)}\le C B_g$, where $C=C(s_g,k,R)$ is independent of $r_0$, such that
\[
    f(x)\;=\;g\bigl(Ax+\Delta h(x)\bigr)
    \qquad\text{for all }x\in T_{r_0,\tau}.
\]
Consequently, for data distributions supported in $T_{r_0,\tau}$, the target belongs to the class $\F_{\sqrt k,\,CB_g,\,\eps}$ of Definition \ref{def:class}, restricted to $T_{r_0,\tau}$, with $\eps$ as in (ii).
\end{enumerate}
\end{proposition}

\begin{proof}[Proof outline; details in Appendix \ref{app:geometry}]
Since $\tau<\rho$, the tube map $(u,v)\mapsto\varphi(u)+\nu(u)v$, with $\nu(u)$ a $C^{s_h}$ orthonormal frame of the normal space $N_{\varphi(u)}M$ and $\norm{v}\le\tau$, is a $C^{s_h}$ diffeomorphism onto a neighborhood of $\varphi(B(0,r_0))$ in the tube, and the chart-coordinate function $x\mapsto u(\pi(x))$ is $C^{s_h}$ there; one derivative is spent inverting the projection, which is why Assumption \ref{assump:geom} requires $C^{s_h+1}$ regularity of $M$.  In the adapted coordinates, $A\varphi(u)=u$, so for $x=\varphi(u)+\nu(u)v$ the pre-correction is exactly
\[
    u(\pi(x))-Ax \;=\; -\,A \nu(u)v .
\]
Rescaling $x\mapsto x/r_0$ reduces all bounds to a manifold that is $\bar\kappa r_0$-close to flat in $C^{s_h+1}$, where every derivative of the pre-correction is $O(\bar\kappa r_0)$; transferring back gives
\[
    \sup_{T_{r_0,\tau}}\norm{D^m\bigl(u(\pi(x))-Ax\bigr)}
    \;\lesssim\;\bar\kappa\, r_0^{\,2-m},
    \qquad 0\le m\le s_h.
\]
Multiplying by a smooth cutoff at scale $r_0$, whose derivatives contribute matching factors of $r_0^{-1}$, and integrating over the support volume $\lesssim r_0^d$ yields \eqref{eq:curvature-bound-safe}.  Property (i) holds because the cutoff equals $1$ on $T_{r_0,\tau}$.  For (iii), let $g$ be a Whitney extension of $f_0\circ\varphi$ from $B(0,r_0)$ to $\R^k$, localized to the bounded feature domain; Whitney's constants are independent of the domain, hence of $r_0$, which is why the chart hypothesis is stated in the uniform norm.  Then $g(h(x))=f_0(\varphi(u(\pi(x))))=f_0(\pi(x))=f(x)$ on the tube, and $h(T_{r_0,\tau})\subset B(0,r_0)$, so the feature domain is bounded as required by Assumption \ref{assump:domain}.
\end{proof}

\begin{remark}[Necessity of the correction]
\label{rem:necessity}
For $\tau>0$ the correction is genuinely required: no purely linear feature map admits the factorization (iii) in general.  Consider $k=1$, $d=2$, $M$ the parabola $\psi(u)=\kappa u^2/2$, and target $f_0(\varphi(u))=u$.  Any linear feature $\ell(x)=\ip{a}{x}$ is constant on parallel lines, whereas the normal fibers of $M$ rotate at rate $\kappa$; hence some fiber within the chart makes an angle $\gtrsim\kappa r_0$ with the level sets of $\ell$, and one can find two points of $T_{r_0,\tau}$ on a common level set of $\ell$ whose chart coordinates differ by $\gtrsim\kappa r_0\tau$.  Any function of $\ell$ is constant on that level set, so $\sup_{T_{r_0,\tau}}|f-g\circ\ell|\gtrsim\kappa r_0\tau$ for every $g$ and every linear $\ell$.  The correction constructed above satisfies $\norm{\Delta h}_{L^\infty}\lesssim\bar\kappa r_0\tau$ on the tube, so it is of the minimal necessary size in sup norm.  For $\tau=0$ (data exactly on $M$) the tangent projection with $\Delta h=0$ already factorizes the target, and the content of the proposition reduces to the distortion statement in (i); the interesting regime for the correction mechanism is noisy or off-manifold data.
\end{remark}

\begin{remark}[Compatibility with Theorem \ref{thm:covering}]
\label{rem:compatibility}
The correction produced by Proposition \ref{prop:geometry} lies in the Sobolev ball $W^{s_h,2}$ with the same smoothness index $s_h>d/2$ required by Assumption \ref{assump:smooth}, so Theorem \ref{thm:covering} applies verbatim with $B_A=\sqrt k$ and $\eps\le C\bar\kappa\, r_0^{\,d/2}\max(r_0^{2-s_h},r_0^2)$: the chain
\[
    \bar\kappa\ \longrightarrow\ \eps\ \longrightarrow\
    \log N(\delta,\F,L^2)\ \longrightarrow\ \text{generalization}
\]
composes in every ambient dimension.  The price is that $M$ must be $C^{s_h+1}$ with controlled derivatives up to order $s_h+1$; only $\kappa_2$ is curvature in the classical sense, and the honest statement is that \emph{curvature at scale $r_0$}—the aggregate $\bar\kappa$—controls the correction.  A $C^2$ formulation is recovered as the case $s_h=2$, which satisfies $s_h>d/2$ only for $d\le3$.
\end{remark}

\begin{remark}[The representation dimension is the manifold dimension]
\label{rem:kdim}
In Definition \ref{def:class} the integer $k$ is the dimension of the representation, a modeling choice.  In Proposition \ref{prop:geometry} it is fixed by the geometry: $k=\dim M$, and $A$ is a tangent chart of rank $k$.  In particular $k$ is the intrinsic dimension of $M$, not the dimension of the smallest linear subspace containing $M$.  If a $5$-dimensional manifold lies in a $20$-dimensional subspace of $\R^d$, the construction uses a rank-$5$ tangent chart and a correction whose size is governed by the curvature of $M$ as a $5$-manifold; the subspace dimension never enters, and the intrinsic term of Theorem \ref{thm:covering} is $5$-dimensional.  Taking $k$ larger than $\dim M$ is admissible, since a chart padded with zero rows is still a member of the matrix ball, but wasteful: the intrinsic entropy term grows as $(B_g/\delta)^{k/s_g}$.  The two notions of dimension coincide only when $M$ is an affine subspace, where $\bar\kappa=0$, $\eps=0$, and the class reduces to a multi-index model.  Thus small $\eps$ does not mean that the data nearly lie on a $k$-dimensional linear subspace; it means that, at the scale $r_0$, the manifold is well approximated by its tangent spaces, which is the local statement that curvature controls.
\end{remark}

\begin{remark}[Choice of localization scale]
\label{rem:scale}
The exponent $d/2+2-s_h$ in (ii) changes sign at $s_h=d/2+2$.  For moderate smoothness, $d/2<s_h<d/2+2$, the bound improves as $r_0\to0$: localizing shrinks the correction faster than the cutoff derivatives inflate it.  For $s_h>d/2+2$, aggressive localization is counterproductive at top derivative order.  This tradeoff becomes quantitative in the atlas construction of Corollary \ref{cor:atlas}, where smaller $r_0$ also increases the number of charts.
\end{remark}

\subsection{From one chart to an atlas}
\label{sec:atlas}

Proposition \ref{prop:geometry} is a single-chart statement.  For data spread over a compact manifold, the construction globalizes by patching charts with a partition of unity; the price is a factor counting the charts.

\begin{definition}[Atlas class]
\label{def:atlas}
Let $M$ be compact and let $\bar x_1,\ldots,\bar x_J\in M$ be centers at which Assumption \ref{assump:geom} holds at scale $r_0$, with local tubes $T^j$ as in Proposition \ref{prop:geometry} covering $M^\tau:=\{x:\mathrm{dist}(x,M)\le\tau\}$, and let $\{\rho_j\}_{j\le J}$ be a fixed $C^{s_h}$ partition of unity subordinate to this cover.  The \emph{atlas class} is
\[
    \F^{\mathrm{atlas}}
    :=\Bigl\{\,\textstyle\sum_{j=1}^J\rho_j\cdot(g_j\circ h_j)\;:\;
    g_j\circ h_j\in\F_{\sqrt k,\,CB_g,\,\eps}\ \text{for all }j\,\Bigr\}.
\]
The cover and partition of unity are fixed by the geometry; the free parameters are the $J$ local models $(g_j,h_j)$.
\end{definition}

\begin{corollary}[Atlas extension]
\label{cor:atlas}
Suppose $M$ is compact with diameter $D$, and the bounds of Assumption \ref{assump:geom} hold at every $\bar x\in M$ at scale $r_0$ with a uniform aggregate $\bar\kappa$ and reach $\rho$; assume $r_0\le\rho$ and fix $0\le\tau\le\min(\rho/2,r_0)$.  Then there exist $J\le C(d,k)\,(D/r_0)^k(D/\rho)^{d-k}$ centers (Remark \ref{rem:chartcount}; the reach factor is a constant when $\rho\gtrsim D$, as for a sphere) $\bar x_1,\ldots,\bar x_J\in M$ whose local tubes $T^j$ (as in Proposition \ref{prop:geometry}, centered at $\bar x_j$) cover the full tube $M^\tau:=\{x:\mathrm{dist}(x,M)\le\tau\}$ with overlap bounded by a constant $L=L(k)$, a fixed $C^{s_h}$ partition of unity $\{\rho_j\}$ subordinate to the cover, and local maps $h_j=A_jx+\Delta h_j$ as in Proposition \ref{prop:geometry} with $\eps_j\le C\bar\kappa\,r_0^{\,d/2}\max(r_0^{\,2-s_h},r_0^{\,2})$ for every $j$.  Any target $f=f_0\circ\pi$ with $\sup_j\norm{f_0\circ\varphi_j}_{C^{s_g}}\le B_g$ satisfies
\[
    f(x)\;=\;\sum_{j=1}^J \rho_j(x)\, g_j\bigl(h_j(x)\bigr)
    \qquad\text{on } M^\tau,
\]
with $\norm{g_j}_{W^{s_g,2}}\le CB_g$, and the atlas class $\F^{\mathrm{atlas}}$ of Definition \ref{def:atlas} built on this cover satisfies
\[
    \log N\bigl(\delta,\F^{\mathrm{atlas}},L_n^2\bigr)
    \;\le\;
    \log N\bigl(\delta,\F^{\mathrm{atlas}},\norm{\cdot}_\infty\bigr)
    \;\lesssim\; J\,\log N\bigl(\delta,\;\F_{\sqrt k,\,CB_g,\,\eps},\;\norm{\cdot}_\infty\bigr),
\]
and by Corollary \ref{cor:supnorm} the right-hand side obeys the bound of Theorem \ref{thm:covering}.
\end{corollary}

\begin{proof}
Take a maximal $r_0/4$-separated set of centers on $M$.  The hypothesis $r_0\le\rho$ ensures that ambient-close points of $M$ lie on a common chart.  Indeed, suppose $\norm{m-\bar x_j}\le r_0/4$ and write $m=(u_m,w_m)$ in the adapted coordinates at $\bar x_j$, so that $\norm{u_m}\le r_0/4$ and the graph point $q:=\varphi_j(u_m)$ is defined.  The vector $v:=m-q=(0,\,w_m-\psi_j(u_m))$ is vertical, while the tangent space $T_qM$ has slope $\norm{D\psi_j(u_m)}\le\bar\kappa r_0\le c_0$, so $\mathrm{dist}(v,T_qM)\ge(1-c_0)\norm v$; Federer's tangent inequality \citep[Thm.~4.18]{federer1959curvature} gives $\mathrm{dist}(v,T_qM)\le\norm v^2/(2\rho)$, whence $v=0$ or $\norm v\ge2\rho(1-c_0)\ge2r_0(1-c_0)\ge\tfrac95\,r_0$ for $c_0\le\tfrac1{10}$.  Since $\norm v\le\norm{w_m}+\norm{\psi_j(u_m)}\le r_0/4+c_0r_0/32<r_0/3$, we conclude $m=q=\varphi_j(u_m)\in\varphi_j(B(0,r_0/2))$.  Maximality of the separated set therefore makes the half-patches $\varphi_j(B(0,r_0/2))$ cover $M$; the packing and tube-volume argument of Remark \ref{rem:chartcount} bounds $J$ and the overlap $L(k)$, and Proposition \ref{prop:geometry} applies at each center.  The factorization identity holds because $\sum_j\rho_j=1$ on $M^\tau$ and $g_j\circ h_j=f$ on the support of $\rho_j$.  For the entropy bound, cover each local class at \emph{sup-norm} accuracy $\delta$ using Corollary \ref{cor:supnorm}.  If $f=\sum_j\rho_j\,(g_j\circ h_j)$ and each $g_j\circ h_j$ is replaced by a net element within sup-distance $\delta$, then pointwise
\[
\Bigl|\,f(x)-\textstyle\sum_j\rho_j(x)\,\widetilde{(g_j\circ h_j)}(x)\Bigr|
\;\le\;\sum_j\rho_j(x)\,\delta\;=\;\delta,
\]
since the $\rho_j$ are nonnegative and sum to one; no factor of $J$ or of the overlap is lost.  The joint net has product structure over the $J$ charts, so the log-covering numbers add.  (The uniform norm is essential here: per-chart \emph{empirical-$L^2$} accuracy $\delta$ would control the combination only up to $\sqrt J\delta$.)
\end{proof}

\begin{remark}[The chart count and its constants]
\label{rem:chartcount}
The bound on $J$ is a packing argument, and since $J$ multiplies every term of the risk bound it is worth recording what the constants depend on.  Let $\bar x_1,\ldots,\bar x_J$ be a maximal $r_0/4$-separated subset of $M$ in the ambient norm.
\begin{enumerate}
\item[(i)] \emph{(Disjoint patches.)}  In the adapted coordinates at $\bar x_j$, the graph patch $\Pi_j:=\varphi_j(B(0,r_0/10))$ has intrinsic volume at least $\omega_k(r_0/10)^k$, a graph having at least the volume of its base, and ambient radius at most $(1+c_0)r_0/10<r_0/8$ because $\norm{D\psi_j}\le c_0$ on the chart.  The $\Pi_j$ are therefore pairwise disjoint, and $J\,\omega_k(r_0/10)^k\le\mathrm{vol}_k(M)$.
\item[(ii)] \emph{(Volume from the reach.)}  The normal exponential map $(y,v)\mapsto y+v$, $v\in N_yM$, $\norm v<\rho$, is injective by the definition of the reach, and its Jacobian $\det(I-\mathrm{II}_y(v))$ is at least $(1-\norm v/\rho)^k$ because the principal curvatures of $M$ are bounded by $1/\rho$.  Integrating over $\norm v\le\rho/2$ gives $\mathrm{vol}_d(M^{\rho/2})\ge2^{-k}\omega_{d-k}(\rho/2)^{d-k}\,\mathrm{vol}_k(M)$, while $M^{\rho/2}$ lies in a ball of radius $D+\rho/2$.  Since $\rho\le D$ for a closed manifold (at a point of $M$ farthest from a fixed point of $M$, some principal curvature is at least $1/D$), this gives $\mathrm{vol}_k(M)\le C(d,k)\,D^k(D/\rho)^{d-k}$.
\item[(iii)] \emph{(The count.)}  Hence
\[
    J\;\le\;C(d,k)\Bigl(\frac{D}{r_0}\Bigr)^{k}\Bigl(\frac{D}{\rho}\Bigr)^{d-k}.
\]
The factor $(D/\rho)^{d-k}$ counts how many sheets of $M$ can pass through an ambient ball.  It is a constant when $\rho\gtrsim D$, as for a sphere, and it is independent of $r_0$ and $n$, so it affects only the constants in Corollary \ref{cor:rates} and Remark \ref{rem:r0choice}.  It is not removable in general: a $k$-manifold coiled through a ball of diameter $D$ with inter-sheet spacing of order $\rho$ has volume of order $D^k(D/\rho)^{d-k}$.
\item[(iv)] \emph{(Overlap.)}  A point $x\in M^\tau$ lies in $T^j$ only if $\norm{\pi(x)-\bar x_j}\le(1+c_0)r_0/2$.  By the common-chart argument in the proof above, run at radius $\tfrac7{10}r_0$ instead of $r_0/4$, the set $M\cap B(\pi(x),\tfrac7{10}r_0)$ is a single graph patch, of volume at most $(1+c_0^2)^{k/2}\omega_k(\tfrac7{10}r_0)^k$, and it contains the disjoint patches $\Pi_j$ of every such center; so their number is at most $L(k):=2\cdot7^k$.
\end{enumerate}
\end{remark}

Combining Corollary \ref{cor:atlas} with the per-chart bound on $\eps$ makes the role of the scale explicit: shrinking $r_0$ reduces the ambient correction term inside each chart while multiplying the number of charts as $r_0^{-k}$.  Section \ref{sec:risk} converts these bounds into an excess-risk theorem and carries out the scale analysis at the level of rates.

\begin{remark}[What the reach hypothesis is used for]
\label{rem:reach}
It is worth recording exactly where the global lower bound on the reach enters the geometric results, since the answer differs between the single-chart statement and the atlas.  In the proof of Proposition \ref{prop:geometry} the reach is used only through the following per-chart consequence, a quantitative form of the clause ``near $\bar x$ the manifold is the graph of $\varphi$'' in Assumption \ref{assump:geom}:
\begin{equation}
    \tag{S}
    \label{eq:solitariness}
    \text{every } y\in M \text{ with }
    \mathrm{dist}\bigl(y,\varphi(B(0,r_0/2))\bigr)\le 2\tau
    \text{ satisfies } y\in\varphi\bigl(B(0,r_0)\bigr),
\end{equation}
i.e.\ no second sheet of $M$ enters the $2\tau$-neighborhood of the chart patch.  Indeed, the quantitative bounds of Appendix \ref{app:geometry} rest on the inverse function theorem applied to the tube map, a perturbation-of-identity argument that uses only $\bar\kappa r_0\le c_0$; the reach appears there only in identifying the local fiber base with the global nearest-point projection (Steps 0--1), and \eqref{eq:solitariness} suffices for this: any nearest-point competitor for $x\in T_{r_0,\tau}$ lies within $2\tau$ of the patch, hence on the graph by \eqref{eq:solitariness}, where injectivity of the tube map settles the comparison.  Proposition \ref{prop:geometry} therefore holds with ``reach at least $\rho$'' and $\tau\le\min(\rho/2,r_0)$ replaced by \eqref{eq:solitariness} and $\tau\le r_0/5$, with the tube-map statements read over the half-patch that the construction actually uses.  Conversely, \eqref{eq:solitariness} follows from Assumption \ref{assump:geom} with $\tau\le\min(\rho/2,\,r_0/5)$, \emph{provided} the graph clause is read with quantified ambient thickness --- $M$ coincides with the graph on the cylinder $B_k(0,r_0)\times B_{d-k}(0,r_0)$: interior nearest-point competitors are excluded by Federer's tangent inequality $\mathrm{dist}(y-p,T_pM)\le\norm{y-p}^2/(2\rho)$ \citep[Thm.~4.18]{federer1959curvature} applied to the normal vector $y-p$, and boundary competitors land inside that cylinder.  As stated, the clause is a germ condition and the reach hypothesis silently supplies the missing thickness; making it explicit is part of the point of \eqref{eq:solitariness}.  Thus \eqref{eq:solitariness} is implied by, but strictly weaker than, a global reach bound: it is unaffected by bottlenecks of $M$ far from the chart.  This matches the dichotomy of \citet[Thm.~3.4]{aamari2019reach}, in the sense that the reach is always attained either by a radius of curvature or at a bottleneck pair: the curvature mechanism is already subsumed by $\bar\kappa r_0\le c_0$, and \eqref{eq:solitariness} is the localized form of the bottleneck mechanism --- a lower bound on the size of the neighborhood on which the curvature-driven local statement is globally valid.  By contrast, the reach performs genuinely global work in Corollary \ref{cor:atlas} that \eqref{eq:solitariness} does not replace: it bounds the volume of $M$, hence the chart count $J\lesssim(D/r_0)^k$; it controls the ambient overlap of the tubes; and it is invoked to ensure that ambient-close points of $M$ lie on a common sheet, so that the charts of a maximal separated set cover $M$.  A manifold satisfying \eqref{eq:solitariness} at every center may still coil through a ball of diameter $D$ with inter-sheet spacing just above $2\tau$, inflating the chart count by a factor of order $(D/\tau)^{d-k}$.  A fully localized atlas would need substitutes for these global inputs (an intrinsic volume bound and an ambient multiplicity bound); we do not pursue this here.  We retain the reach formulation throughout for continuity with the manifold-learning literature \citep{niyogi2008finding}; the decomposition matters because \eqref{eq:solitariness} is per-chart, the form that admits gauge-dependent analogues in model geometries carrying no invariant metric, where a global reach is not even definable.
\end{remark}

\section{From entropy and geometry to sample complexity}
\label{sec:risk}

We now combine Theorem \ref{thm:covering} with the geometric results of Section 4 into a finite-sample excess-risk bound, so that the geometric and entropic results compose into a single risk bound.

\paragraph{Setting.}
Let $(x_1,y_1),\ldots,(x_n,y_n)$ be i.i.d.\ with $x\sim P$ supported on the tube $M^\tau$ of Corollary \ref{cor:atlas}, $\E[y\mid x]=f^*(x)$, and $|y|\le b$ almost surely, where $f^*=f_0\circ\pi$ satisfies the hypotheses of Corollary \ref{cor:atlas}, so that $f^*\in\F^{\mathrm{atlas}}$; assume also that the members of $\F^{\mathrm{atlas}}$ are uniformly bounded by $b$ (Assumption \ref{assump:domain}).  Fix a net accuracy $\delta>0$ and let $G_\delta\subset\F^{\mathrm{atlas}}$ be an \emph{internal} $\delta$-net of minimal cardinality: its points belong to $\F^{\mathrm{atlas}}$, so they share the uniform bound $b$, and $|G_\delta|\le N(\delta/2,\F^{\mathrm{atlas}},\norm{\cdot}_\infty)$, since each ball of an external $\delta/2$-net meeting the class can be recentered at a member (the cardinality is then bounded by Corollaries \ref{cor:supnorm} and \ref{cor:atlas}).  Define the \emph{sieve least-squares estimator}
\[
    \hat f\;:=\;\arg\min_{f\in G_\delta}\ \frac1n\sum_{i=1}^n\bigl(f(x_i)-y_i\bigr)^2 .
\]
This is a sieve estimator in Grenander's sense: $G_\delta$ is a finite approximating subclass whose resolution $\delta$ is tuned to the sample size.  It is a device of proof rather than an algorithm---the net is not constructive and $|G_\delta|$ is exponential in the entropy---and the penalized estimator of Proposition \ref{prop:representer} is not analyzed here.  Sieve estimators of this type are classical in nonparametric regression \citep{yang1999information,vandervaart2008rates}; we use one because it admits an elementary and fully rigorous analysis over the nonconvex class $\F^{\mathrm{atlas}}$, avoiding the localization technology that empirical risk minimization over nonconvex classes would require.

\begin{theorem}[Excess risk over the atlas class]
\label{thm:risk}
In the setting above there is a constant $C=C(b)$ such that for every $\delta>0$ and $\eta\in(0,1)$, with probability at least $1-\eta$,
\[
    \norm{\hat f-f^*}_{L^2(P)}^2
    \;\le\;
    C\Bigl(\delta^2
    +\frac{\log N\bigl(\delta/2,\F^{\mathrm{atlas}},\norm{\cdot}_\infty\bigr)+\log(2/\eta)}{n}\Bigr).
\]
Optimizing over $\delta$ with the entropy bound of Corollary \ref{cor:atlas} yields, with probability at least $1-\eta$ and up to constants depending on $b$, $B_g$, and the parameters of the assumptions,
\begin{equation}
    \norm{\hat f-f^*}_{L^2(P)}^2
    \;\lesssim\;
    \frac{J\,kd\log n+\log(2/\eta)}{n}
    \;+\;\Bigl(\frac{J}{n}\Bigr)^{\!\frac{2s_g}{2s_g+k}}
    \;+\;\eps^{\frac{2d}{2s_h+d}}\Bigl(\frac{J}{n}\Bigr)^{\!\frac{2s_h}{2s_h+d}},
    \label{eq:risk-bound}
\end{equation}
where $J\lesssim(D/r_0)^k(D/\rho)^{d-k}$ is the chart count (Remark \ref{rem:chartcount}) and $\eps$ the per-chart correction radius.
\end{theorem}

The proof, given in full in Appendix \ref{app:risk}, is a Bernstein bound for excess-loss differences with the variance--excess-risk link, a union bound over the net, and a bias term from the net accuracy.  The three terms of \eqref{eq:risk-bound} are the parametric cost of the linear maps, the intrinsic $k$-dimensional nonparametric rate, and the ambient contribution, which enters only through $\eps$.

\begin{corollary}[Curvature-driven rates]
\label{cor:rates}
Under the hypotheses of Corollary \ref{cor:atlas}, $\eps\le C\bar\kappa\,r_0^{\,d/2+2-s_h}$ for $r_0\le1$ and integer $s_h\ge2$, so \eqref{eq:risk-bound} becomes
\begin{align}
    \norm{\hat f-f^*}_{L^2(P)}^2
    \;&\lesssim\;
    \frac{(D/r_0)^{k}\,kd\log n+\log(2/\eta)}{n}
    \;+\;\Bigl(\frac{D}{r_0}\Bigr)^{\!\frac{2ks_g}{2s_g+k}} n^{-\frac{2s_g}{2s_g+k}}
    \nonumber\\
    &\quad+\;\bar\kappa^{\frac{2d}{2s_h+d}}
    \Bigl(\frac{D}{r_0}\Bigr)^{\!\frac{2ks_h}{2s_h+d}}
    r_0^{\,(d/2+2-s_h)\frac{2d}{2s_h+d}}\,
    n^{-\frac{2s_h}{2s_h+d}} .
    \label{eq:rates}
\end{align}
where the constants absorb the reach factor $(D/\rho)^{d-k}$ of Remark \ref{rem:chartcount}, which depends on neither $r_0$ nor $n$.  In particular, at any fixed scale $r_0$, the ambient term vanishes as $\bar\kappa\to0$ and the rate reduces to the parametric-plus-intrinsic rate
\[
    \Bigl(\frac{D}{r_0}\Bigr)^{\!k}\frac{kd\log n}{n}
    \;+\;C(r_0)\,n^{-\frac{2s_g}{2s_g+k}} ,
\]
the curvature-to-risk statement in its final form.
\end{corollary}

\begin{remark}[Choice of the localization scale]
\label{rem:r0choice}
Collect the $r_0$-exponents in \eqref{eq:rates}.  The parametric and intrinsic terms grow as $r_0$ decreases (through the chart count), while the ambient term scales as $r_0^{\,e_h}$ with
\[
    e_h \;=\; \frac{d(d+4)-2s_h(d+k)}{2s_h+d}.
\]
If $e_h\le0$, i.e.\ $s_h\ge d(d+4)/(2(d+k))$, then all three terms are nonincreasing in $r_0$, and the optimal choice is the largest scale the geometry admits, $r_0\asymp\min(1,\,c_0/\bar\kappa)$ subject to $r_0\le\rho$ and $\tau\le\min(\rho/2,r_0)$: localization finer than the curvature requires is never helpful.  Note that for $k\ge4$ this case always obtains, since then $d(d+4)/(2(d+k))\le d/2<s_h$.  If $e_h>0$ (possible only for $k\le3$), the ambient term improves under localization, and---whenever the parametric term is of lower order---the optimal scale balances the ambient term against the intrinsic term:
\[
    r_0^*\;\asymp\;\Bigl(\bar\kappa^{-\frac{2d}{2s_h+d}}\,n^{-(a-b)}\Bigr)^{\!\frac{1}{e_h+ka}},
    \qquad
    a=\frac{2s_g}{2s_g+k},\quad b=\frac{2s_h}{2s_h+d},
\]
absorbing diameter factors into the constant and truncating to the admissible range; the resulting rate is the intrinsic rate inflated by the chart-count factor $(D/r_0^*)^{ka}$.  Comparing with Remark \ref{rem:scale}: the single-chart threshold $s_h<d/2+2$ is lowered to $s_h<d(d+4)/(2(d+k))$ here---the localization-helps regime shrinks---because chart multiplicity penalizes localization.
\end{remark}

\section{A wrong-geometry separation}
\label{sec:separation}

The results so far are upper bounds: small curvature yields a small correction, and a small correction yields low complexity and fast rates.  This section proves a converse about the choice of the prior itself, first for a one-dimensional model problem and then, for representations, in any dimension: on data around a \emph{curved} manifold, the affine compositional prior provably underperforms the matching curved model class.  The correction budget is thus necessary, not merely sufficient---and in $L^2(P)$, strengthening the sup-norm statement of Remark \ref{rem:necessity}.

\paragraph{Setting.}
This section uses a single-chart model problem, distinct from the atlas setting of Section \ref{sec:risk}.  Let $M=S^1(R)\subset\R^2$ be the circle of radius $R$, with curvature $\kappa=1/R$; in this section $R$ denotes the circle radius, and it plays the role of the domain bound of Assumption \ref{assump:domain} up to the factor $3/2$, since $\sup_{x\in T}\norm{x}\le R+\tau\le3R/2$.  Fix a chart arc of length $r_0$ with $\kappa r_0\le c_0$, where $c_0$ is now an absolute constant (the case $d=2$, $k=1$ of Proposition \ref{prop:geometry}; the arguments below use only $c_0\le1/10$), and a tube radius $0<\tau\le r_0$ with $\tau\le R/2$.  Write points of the tube as
\[
    x=\varphi(u)+v\,n(u),
    \qquad u\in[-r_0/2,\,r_0/2],\quad v\in[-\tau,\tau],
\]
where $\varphi$ is the arclength parameterization of the chart arc and $n(u)$ the outward unit normal (the radial direction); let $T\subset\R^2$ denote this tube piece.  The coordinate map has Jacobian $1+O(\kappa\tau)=1+O(c_0)$.  Let $\pi$ be the radial nearest-point projection onto $M$ and let the target be the chart coordinate transported along fibers,
\[
    f(x):=u(\pi(x)),
\]
the canonical unit-intrinsic-gradient target.  Let $P$ be any data distribution on $T$ with density bounded below by $c_P^{-1}$ times the uniform density on $T$; all constants below depend on $c_P$, and the proofs reduce to $P=\mathrm{Unif}(T)$.  Throughout this section, $\eps_\infty$ denotes a \emph{sup-norm} bound on the correction, to be distinguished from the Sobolev radius $\eps$ of Definition \ref{def:class}; the two are linked by the embedding $\norm{\Delta h}_{L^\infty}\le C_E\eps$ of Appendix \ref{app:covering}, Bookkeeping (ii).

\paragraph{The two classes.}
The \emph{affine compositional class} consists of predictors
\[
    F(x)=g\bigl(\ip{a}{x}+\Delta h(x)\bigr),
    \qquad a\in\R^2,\quad
    \norm{\Delta h}_{L^\infty(T)}\le\eps_\infty,\quad
    \mathrm{Lip}(g)\le L_g ,
\]
with $g:\R\to\R$ arbitrary subject to the Lipschitz bound.  (An affine offset in the feature is absorbed into $g$; for ambient dimension $d>2$ with the circle embedded in a coordinate plane, project $a$ onto the plane---the orthogonal component is constant on $T$.  The vector $a$ is \emph{unconstrained}: the floor below does not degrade for large or small $\norm{a}$.)  The \emph{circle-model class} consists of predictors
\[
    G(x)=g\bigl(u_\theta(x)\bigr),
    \qquad u_\theta(x):=R\,\arg(x)+R\theta,\quad
    \theta\in(-\pi,\pi],\quad
    \norm{g}_{W^{s_g,2}}\le B_g,
\]
where $\arg$ is a fixed continuous branch of the angle on the arc-tube $T$ (one exists because $T$ omits a half-plane): the features are the arclength charts of the radial projection, parameterized by the rotation group, with $\norm{u_\theta-u_{\theta'}}_{L^\infty(T)}=R|\theta-\theta'|$.

\begin{lemma}[Near-constancy on level sets]
\label{lem:constancy}
Let $F(x)=g(\ip{a}{x}+\Delta h(x))$ be a member of the affine class.
For every $t\in\R$, the oscillation of $F$ on the level set
$\{x\in T:\ip{a}{x}=t\}$ is at most $\eta:=2L_g\eps_\infty$.
\end{lemma}

\begin{proof}
On the level set the feature $\ip{a}{x}+\Delta h(x)$ ranges in an
interval of length at most $2\eps_\infty$, and $g$ is
$L_g$-Lipschitz.
\end{proof}

\begin{lemma}[Disintegration bound]
\label{lem:disint}
Let $\mu$ be the law of $\ip{a}{X}$ under $P$ and let
$\mathrm{sd}(f\mid t)$ denote the conditional standard deviation of
$f(X)$ given $\ip{a}{X}=t$.  Then for every $F$ in the affine class,
\[
    \E_P\bigl[(F-f)^2\bigr]
    \;\ge\;\int\bigl(\mathrm{sd}(f\mid t)-\eta\bigr)_+^2\,d\mu(t).
\]
\end{lemma}

\begin{proof}
Condition on $\ip{a}{X}=t$ and let $c_t:=\E[F\mid t]$.  By Lemma
\ref{lem:constancy}, $|F-c_t|\le\eta$ on the level set, so
\[
    \norm{F-f}_{L^2(\cdot\mid t)}
    \ge\norm{f-c_t}_{L^2(\cdot\mid t)}-\norm{F-c_t}_{L^2(\cdot\mid t)}
    \ge\inf_c\norm{f-c}_{L^2(\cdot\mid t)}-\eta
    =\mathrm{sd}(f\mid t)-\eta .
\]
Square, take the positive part, and integrate over $t$.  (For $P$
uniform on a compact planar region and conditioning on a linear
functional, the disintegration is Fubini in the rotated frame: the
conditional law given $t$ is normalized one-dimensional Lebesgue
measure on the chord, and $t\mapsto\mathrm{sd}(f\mid t)$ is
measurable.  The same applies verbatim to restrictions of $P$ to
measurable subsets, which is how the lemma is used below.)
\end{proof}

\begin{lemma}[Shear lemma]
\label{lem:shear}
There is an absolute constant $c>0$ such that for every $a\ne0$ the
following holds.  There is a measurable sub-tube $T'\subset T$ with
$P(T')\ge1/5$ such that, writing $\mu'$ for the law of $\ip{a}{X}$
under $P|_{T'}$ and $\mathrm{sd}'(f\mid t)$ for the conditional
standard deviation under $P|_{T'}$,
\[
    \mu'\bigl\{t:\ \mathrm{sd}'(f\mid t)\ge c\,\kappa r_0\tau\bigr\}
    \;\ge\;\tfrac25 .
\]
For $a=0$ the trivial level set is all of $T$ (take $T'=T$), on
which $\mathrm{sd}(f)\ge cr_0\ge c\kappa r_0\tau$.
\end{lemma}

The proof, an explicit transversality and mass-accounting argument
with all constants displayed, is given in Appendix
\ref{app:separation}.

\begin{theorem}[Wrong-geometry floor]
\label{thm:floor}
There are constants $c,C>0$, depending only on the comparability
constant $c_P$ of the data distribution, such that every member $F$
of the affine class satisfies
\[
    \E_P\bigl[(F-f)^2\bigr]
    \;\ge\;
    c\,\bigl(\kappa r_0\tau- CL_g\eps_\infty\bigr)_+^{2}.
\]
In particular, for the Sobolev class $\F_{B_A,B_g,\eps}$ of
Definition \ref{def:class}, instantiated with $k=1$ and $d=2$, using
$L_g\lesssim B_g$ (valid since $s_g>k/2+1=3/2$ by Assumption
\ref{assump:smooth}) and
$\norm{\Delta h}_{L^\infty}\le C_E\eps$,
\[
    \inf_{F\in\F_{B_A,B_g,\eps}}
    \E_P\bigl[(F-f)^2\bigr]
    \;\ge\;
    c\,\bigl(\kappa r_0\tau-CB_g\eps\bigr)_+^{2},
\]
which is a nonvanishing approximation floor whenever
$\eps\le c'\kappa r_0\tau/B_g$.
\end{theorem}

\begin{proof}
Reduce to $P=\mathrm{Unif}(T)$ at the cost of the factor $c_P$.
Since the integrand is nonnegative,
\[
    \E_P\bigl[(F-f)^2\bigr]
    \;\ge\;P(T')\,\E_{P|_{T'}}\bigl[(F-f)^2\bigr],
\]
with $T'$ the sub-tube of Lemma \ref{lem:shear}.  Apply Lemma
\ref{lem:disint} under $P|_{T'}$ (its conditional laws are the
uniform measures on $\text{line}\cap T'$, and Lemma
\ref{lem:constancy} holds on any subset of a level set), then Lemma
\ref{lem:shear}:
\[
    \E_P\bigl[(F-f)^2\bigr]
    \;\ge\;\tfrac15\cdot\tfrac25\,
    \bigl(c\,\kappa r_0\tau-2L_g\eps_\infty\bigr)_+^2 .
\qedhere
\]
\end{proof}

\begin{corollary}[The floor-or-capacity dilemma]
\label{cor:dilemma}
On the curved-data problem above, the affine compositional prior
faces a dilemma in its correction budget $\eps$:
\begin{enumerate}
\item[(a)] if $\eps\le c'\kappa r_0\tau/B_g$, every predictor in
$\F_{B_A,B_g,\eps}$ incurs squared error at least
$c(\kappa r_0\tau)^2/4$, regardless of sample size;
\item[(b)] if $\eps>c'\kappa r_0\tau/B_g$, then by Remark
\ref{rem:lower} the class contains a $d$-dimensional Sobolev ball of
radius proportional to $\eps$, and its metric entropy carries the
ambient term $(c\eps/\delta)^{d/s_h}$.
\end{enumerate}
Note that the two horns are of different natures: horn (a) is a
genuine risk floor for \emph{every} predictor in the class, whereas
horn (b) constrains only what the prior can \emph{certify}---a
large-$\eps$ class still contains near-perfect predictors on this
problem (Proposition \ref{prop:geometry} constructs one), but its
capacity bound is then ambient-dimensional.  The honest summary is
that the affine prior cannot simultaneously certify small capacity
and small approximation error on curved data.
\end{corollary}

\begin{proposition}[Circle-model upper bound]
\label{prop:circle-model}
Suppose $B_g\ge C$ for a constant depending only on $r_0$ and the
chart conventions, and assume as before that the class members and
$|y|$ are bounded by $b$.  Then:
\begin{enumerate}
\item[(i)] \emph{(Zero correction.)}  The target satisfies
$f=g^*\circ u_{\theta_0}$ exactly on $T$, for the chart-aligned
rotation $\theta_0$ and an outer function $g^*$ (the identity on the
chart interval, smoothly localized) with
$\norm{g^*}_{W^{s_g,2}}\le C\le B_g$: the circle-model class
contains $f$ with correction identically zero.
\item[(ii)] \emph{(Entropy.)}
$\log N\bigl(\delta,\{g\circ u_\theta\},\norm{\cdot}_\infty\bigr)
\lesssim\log\bigl(CB_gR/\delta\bigr)
+\bigl(CB_g/\delta\bigr)^{1/s_g}$:
a one-dimensional parametric term for the rotation group and the
intrinsic ($k=1$) Sobolev term.  No ambient term appears.
\item[(iii)] \emph{(Rate.)}  For i.i.d.\ data on $T$ with
$\E[y\mid x]=f(x)$ and $|y|\le b$, the sieve least-squares estimator
over an internal sup-norm net of the circle-model class attains,
with probability $1-\eta$,
\[
    \norm{\hat f-f}_{L^2(P)}^2
    \;\lesssim\;
    \frac{\log n+\log(2/\eta)}{n}
    \;+\;n^{-\frac{2s_g}{2s_g+1}} .
\]
\end{enumerate}
\end{proposition}

\begin{proof}
(i) is the definition of $u_{\theta_0}$: on the tube,
$u_{\theta_0}(x)=u(\pi(x))=f(x)$ for $\theta_0$ chosen so that the
branch of $\arg$ is centered on the chart, with $g^*$ the identity
on the chart interval, localized by a fixed cutoff to make it
globally Sobolev with norm depending only on the chart length.
(ii): $\norm{g\circ u_\theta-g'\circ u_{\theta'}}_\infty
\le\norm{g-g'}_\infty+L_g\,R\,|\theta-\theta'|$, so a product of a
$\delta/(2L_gR)$-net of the rotation angles (cardinality
$\lesssim B_gR/\delta$) and a sup-norm $\delta/2$-net of the outer
Sobolev ball (log-cardinality $\lesssim(B_g/\delta)^{1/s_g}$, valid
since $s_g>1/2$) covers the class.
(iii): apply Lemma \ref{lem:sieve} with $G$ an internal $\delta$-net
of the circle-model class (its members share the uniform bound $b$,
and $f$ belongs to the class by (i)), then optimize the net accuracy
over the two entropy terms of (ii) exactly as in Appendix
\ref{app:risk}.
\end{proof}

\begin{theorem}[Separation]
\label{thm:separation}
On the curved-data problem of this section with
$\eps\le c'\kappa r_0\tau/B_g$, for every sample size $n$:
\begin{align*}
    \inf_{F\in\F_{B_A,B_g,\eps}}\E_P\bigl[(F-f)^2\bigr]
    &\;\ge\;c\,(\kappa r_0\tau)^2,\\
    \text{while}\qquad
    \norm{\hat f_{\mathrm{circ}}-f}_{L^2(P)}^2
    &\;\lesssim\;
    n^{-\frac{2s_g}{2s_g+1}}+\frac{\log n}{n}
\end{align*}
with high probability, where $\hat f_{\mathrm{circ}}$ is the sieve
estimator over the circle-model class.  In particular, for
$n\gtrsim(\kappa r_0\tau)^{-(2s_g+1)/s_g}$, the matching model class
provably outperforms every predictor in the affine class.
\end{theorem}

\begin{proof}
Combine Theorem \ref{thm:floor} with Proposition
\ref{prop:circle-model}(iii).
\end{proof}

\begin{remark}[What the separation does and does not say]
\label{rem:sep-honesty}
The separation is \emph{class-level}: it lower-bounds the error of
every predictor in the affine compositional class with a constrained
correction budget, not the minimax risk over all estimators.  An
estimator aware of the circle (e.g., the circle-model sieve itself)
is unaffected---indeed that is the point.  The statement quantifies
the price of the \emph{prior}, not of estimation: on curved data the
affine prior either misspecifies or inflates its capacity, while the
matching model prior does neither.  This is the two-sided
justification of $\eps$ as an effective-dimension dial, and the
motivation for model selection over a hierarchy of geometries.
The same caution applies to the ambient estimation
term: it is not forced on curved data, even on estimators that
output members of the affine class---an estimator that knew the
geometry could output the corrected representation of Proposition
\ref{prop:geometry} and learn only the outer function, paying the
intrinsic rate.  What curved data force is the floor; the ambient
term is the price of not knowing which correction to use.
\end{remark}

\begin{remark}[Relation to the earlier results]
\label{rem:sep-relation}
Theorem \ref{thm:floor} upgrades Remark \ref{rem:necessity}
(parabola, sup norm) to an $L^2(P)$ statement under a mild spread
condition on $P$, and Corollary \ref{cor:dilemma} pairs it with
Remark \ref{rem:lower}.  Together with Theorem \ref{thm:risk}, the
mechanism ``the correction size $\eps$ interpolates between
intrinsic and ambient behavior'' now has matching statements on both
sides for the one-dimensional model problem.
\end{remark}

\paragraph{The general case: a representation floor.}
The naive generalization of Theorem \ref{thm:floor} to $S^k(R)$,
$k\ge2$---``for every $A\in\R^{k\times d}$ and every fixed scalar
target the floor $c(\kappa r_0\tau)^2$ holds''---is \emph{false}.
Counterexample ($k=2$, $d=3$): take the chart at the pole $Re_3$ of
$S^2(R)$, target $f=u_1\circ\pi$, and features $Ax=(x_1,x_3)$.  On
the chart, $\ip{\partial\varphi/\partial u_1}{e_2}=O(\kappa^2r_0^2)$
and $x_3$ determines the radial coordinate up to the
$O(\kappa r_0^2)$ sag of the sphere, so the feature pair $(x_1,x_3)$
determines $v$ up to $O(\kappa r_0^2)$ and hence $u_1$ up to error
$O(\kappa^2r_0^3)$; consequently
\[
    g(x_1,x_3)\;:=\;\frac{x_1R}{R+\hat v(x_1,x_3)}
\]
is an $O(1)$-Lipschitz member of the affine class with
$\Delta h\equiv0$ and squared error $O(\kappa^2r_0^3)^2\ll
(\kappa r_0\tau)^2$ whenever $\tau\gg\kappa r_0^2$---a nonempty
regime under the standing hypotheses.  The mechanism: a scalar
target on a chart of dimension $k\ge2$ has a flat direction at every
point, the tangent to its level set, and when that direction is the
same across the chart---as it is for any function of a single linear
chart coordinate, not only for $u_1$---the $k$ features can align
the kernel of $A$ with it and spend the remaining freedom on the
normal coordinate.  For $k=1$ a nondegenerate scalar target has no
flat direction, so the single feature must choose between the chart
coordinate and the normal coordinate; that choice is the content of
the shear lemma.

The correct higher-dimensional statement is therefore not about a
fixed target but about a fixed \emph{representation}: no affine
feature map with a small correction budget serves \emph{all} targets
on a curved chart.  The supremum over targets is taken after the
feature map is fixed, so the target may depend on the kernel of $A$.

\begin{theorem}[Representation floor in any dimension]
\label{thm:repfloor}
Let $M=S^k(R)\subset\R^{k+1}$ with $k\ge1$ and $\kappa=1/R$, let
$p\in M$, let $U\subset M$ be the closed geodesic ball of radius
$r_0/2$ about $p$ with $\kappa r_0\le c_0\le1/10$, and let
\[
    T\;:=\;\bigl\{(1+v/R)\,y:\ y\in U,\ |v|\le\tau\bigr\},
    \qquad 0<\tau\le\min(r_0,R/2),
\]
with $P$ a distribution on $T$ whose density is at least $c_P^{-1}$
times the uniform density.  Let $\mathcal T$ be the class of targets
$f_0\circ\pi$ with $\norm{f_0}_{C^{s_g}(U)}\le B_T$ in geodesic
normal coordinates at $p$, where $B_T=C(k,s_g)\,(r_0+r_0^{1-s_g})$.
There are constants $c,C>0$ depending only on $k$ and $c_P$ such
that for every $A\in\R^{k\times(k+1)}$, every $\Delta h$ with
$\norm{\Delta h}_{L^\infty(T)}\le\eps_\infty$, and every $L_g>0$,
\begin{equation}
    \sup_{f\in\mathcal T}\ \inf_{\mathrm{Lip}(g)\le L_g}\
    \E_P\bigl[\bigl(g(Ax+\Delta h(x))-f(x)\bigr)^2\bigr]
    \;\ge\;c\,\bigl(\kappa r_0\tau-CL_g\eps_\infty\bigr)_+^2 .
    \label{eq:repfloor}
\end{equation}
The supremum is attained, up to the constants, by an explicit
target: fix a unit vector $w\in\ker A$, let $q\in\{Rw,-Rw\}$ be the
point nearer to $p$, let $d_q$ be the geodesic distance from $q$ on
$M$, and set
\[
    f_q\;:=\;\frac{d_q^{\,2}-d_q(p)^2}{2\max\bigl(d_q(p),\,r_0\bigr)}
    \circ\pi .
\]
Consequently, for the Sobolev class of Definition \ref{def:class}
with $d=k+1$ and $B_g\ge B_T$, using $L_g\lesssim B_g$ and
$\norm{\Delta h}_\infty\le C_E\eps$,
\[
    \inf_{h}\ \sup_{f\in\mathcal T}\ \inf_{g}\
    \E_P\bigl[(g\circ h-f)^2\bigr]
    \;\ge\;c\,(\kappa r_0\tau-CB_g\eps)_+^2 ,
\]
the outer infimum over feature maps $h=Ax+\Delta h$ with
$\norm{\Delta h}_{W^{s_h,2}}\le\eps$ and any $A$, the inner one over
$g$ in the Sobolev ball of radius $B_g$.
\end{theorem}

The proof, in Appendix \ref{app:separation}, is the argument of
Theorem \ref{thm:floor} run along the lines parallel to a kernel
direction $w$ of $A$, with the target chosen so that it varies along
those lines everywhere except near the one point $q$ at which $w$ is
normal to the sphere.  The adversarial target is interpretable: when
$q$ is far from the chart it is, to first order, the chart coordinate
along the tangential projection of $w$, which is exactly the
coordinate the counterexample above sacrifices (there $\ker A=\R e_2$
and $q$ lies on the equator in direction $e_2$); when $w$ is normal
at the chart center it is the squared geodesic distance from the
center.  The targets in $\mathcal T$ are those covered by
Proposition \ref{prop:geometry}(iii) with $B_g\ge B_T$, so the pair
of statements reads: one feature map with budget proportional to
curvature serves every target on the chart (Proposition
\ref{prop:geometry}), and no feature map with a smaller budget does
(Theorem \ref{thm:repfloor}).  Two variants follow from the same
proof and are not pursued: vector-valued targets, such as the chart
map itself, which vary in every intrinsic direction and need no
adversarial choice; and fixed scalar targets whose flat direction
rotates across the chart, for which the floor should hold with a
constant depending on the rotation rate.

\begin{remark}[Toward an oracle over geometries]
\label{rem:oracle}
With Theorem \ref{thm:separation} in hand, a model-selection
statement takes shape: given a finite hierarchy of model geometries
with model classes $\F^{(i)}$, a penalized sieve selector over
$\bigcup_i\F^{(i)}$ pays the parametric price of the selected
geometry's symmetry group plus the intrinsic rate, while the floor
guarantees that geometries whose flatness defect on the data exceeds
their budget are eliminated by the data themselves; the two-geometry
case is contained in the results above.  A parallel program exists
on the invariance side: for kernel regression on manifolds, exact
invariance under a group action reduces the effective dimension to
that of the quotient \citep{tahmasebi2023exact}; the correction
budget $\eps$, reinterpreted as a deviation from invariance,
suggests an approximate-invariance interpolation with that result as
its $\eps=0$ endpoint.  We leave both directions to future work; the
following remark makes the class-level statement exact.
\end{remark}

\begin{remark}[The correction budget as an invariance defect]
\label{rem:invariance}
The reinterpretation of $\eps$ proposed in Remark \ref{rem:oracle} is
exact at the level of the class, in the following sense.  Fix
$f=g(Ax+\Delta h(x))\in\F_{B_A,B_g,\eps}$ and write $h(x)=Ax+\Delta h(x)$.
For every translation $t\in\R^d$,
\begin{equation}
    h(x+t)-h(x)-At \;=\; \Delta h(x+t)-\Delta h(x),
    \label{eq:defect-identity}
\end{equation}
so by translation invariance of the Sobolev norm,
\[
    \sup_{t\in\R^d}\norm{h(\cdot+t)-h(\cdot)-At}_{W^{s_h,2}(\R^d)}\;\le\;2\eps .
\]
Thus every feature map in the class is $2\eps$-equivariant under the full
translation group, with linear intertwiner $t\mapsto At$, and the defect is
small at full Sobolev strength, not merely in sup norm.  For $t$ in the
subgroup $\ker A$, which has dimension at least $d-k$, the linear term in
\eqref{eq:defect-identity} vanishes; the embedding
$W^{s_h,2}(\R^d)\hookrightarrow C^0$ (valid since $s_h>d/2$) and the
Lipschitz bound on $g$ (Assumption \ref{assump:smooth}) then give
\[
    \bigl|f(x+t)-f(x)\bigr|
    \;\le\; L_g\,\norm{\Delta h(\cdot+t)-\Delta h}_{\infty}
    \;\le\; C(d,s_h)\,B_g\,\eps
    \qquad\text{for every } t\in\ker A
\]
and every $x$ with $x,x+t\in\X$ (with $\sup$ over all of $\R^d$ if $g$ is
read as globally defined).  Thus each member of the class is
$CB_g\eps$-invariant under a translation subgroup of dimension at least
$d-k$; the class is a union of such families, one for each admissible $A$,
and the parametric term $kd\log(CB_AB_g/\delta)$ in Theorem
\ref{thm:covering} includes, in its Grassmannian factor of $k(d-k)$
parameters, the price of selecting the subgroup (the remaining $k^2$
parameters price the intertwiner).  At $\eps=0$ the invariance is exact and
each member factors through the quotient $\R^d/\ker A$, of dimension
$\operatorname{rank}A\le k$: this is an exact-invariance endpoint at which
effective dimension drops to that of the quotient, in the spirit of the
kernel-on-manifold results of \citet{tahmasebi2023exact}, and for the class
$\F_{B_A,B_g,\eps}$ the covering bound \eqref{eq:covering-bound} is the
corresponding approximate-invariance interpolation: the ambient term
vanishes at the invariant endpoint and grows with the budget $\eps$.  What
remains open is the analogous interpolation in the kernel-on-manifold
setting itself, as discussed in Remark \ref{rem:oracle}.  We caution that
the converse direction is genuinely weaker: near-equivariance in sup norm
constrains only the $C^0$ distance to some affine map (Hyers--Ulam
stability of the Cauchy equation; Skof's version on restricted domains),
with no control at the derivative orders that the budget $\eps$ measures,
so sup-norm near-equivariance does not imply membership in
$\F_{B_A,B_g,\eps}$.  The rigidity question at Sobolev strength is left to
future work.
\end{remark}

\section{The correction budget as an invariance defect}
\label{sec:invariance}

Since \(h(x+t)-h(x)-At=\Delta h(x+t)-\Delta h(x)\), every feature map in the class is \(2\eps\)-equivariant under translations with linear intertwiner \(t\mapsto At\), at full Sobolev strength; for \(t\in\ker A\) (dimension \(\ge d-k\)) the Sobolev embedding gives \(|f(x+t)-f(x)|\le C(d,s_h)B_g\eps\), so each member is \(CB_g\eps\)-\emph{invariant} under a \((d-k)\)-dimensional translation subgroup.  At \(\eps=0\) each member factors through the quotient \(\R^d/\ker A\), of dimension \(\le k\)---the exact-invariance endpoint, in the spirit of \citet{tahmasebi2023exact}---so \eqref{eq:covering-bound} is an approximate-invariance interpolation (Remark \ref{rem:invariance}).  We claim no minimax optimality for the composed class, and the higher-dimensional separation is proved only at the level of representations (Theorem \ref{thm:repfloor}), not for a fixed scalar target.

\section{Related work}

\paragraph{Function priors and nonparametric statistics.}
The view of learning as function estimation under a prior or regularizer is classical in nonparametric statistics and Bayesian nonparametrics \citep{ghosal2000convergence,vandervaart2008rates}.  Metric entropy and prior mass conditions play a central role in posterior contraction theory, while covering numbers and Rademacher complexity play analogous roles in frequentist generalization bounds \citep{dudley1967sizes,yang1999information,bartlett2002rademacher,cucker2002mathematical,wainwright2019high}.

\paragraph{Kernel methods and RKHS regularization.}
Classical RKHS theory provides a precise model of quadratic function-space regularization \citep{aronszajn1950theory,scholkopf2002learning,steinwart2008support,bach2024learning}.  Our work uses RKHS and Sobolev tools for the component classes, but the induced prior on \(f=g\circ h\) is not a single quadratic RKHS penalty.  Representer theorems have also been extended beyond the quadratic RKHS setting to Banach spaces associated with neural network classes \citep{parhi2021banach}; Proposition \ref{prop:representer} is instead a componentwise application of the classical theorem.

\paragraph{Multi-index and feature-learning models.}
Multi-index models study functions of low-dimensional projections, such as \(f(A^Tx)\); classical estimators recover the index space by inverse regression or gradient-based methods \citep{li1991sliced,xia2002adaptive}, with quantitative guarantees depending on the intrinsic dimension under suitable assumptions \citep{plan2016generalized,klock2020estimating}.  Neural networks trained by gradient methods can also learn single- and multi-index structure \citep{bietti2022learning,damian2022neural}.  Our model extends this setting by allowing a controlled nonlinear correction \(\Delta h\), thereby interpolating between a linear projection model and a richer nonlinear representation class.

\paragraph{Compositional regression rates.}
Closest to our statistical bounds is the line of work that assumes the \emph{target} is compositional and derives improved rates.  \citet{schmidt2020nonparametric} shows that deep ReLU networks attain minimax rates governed by the effective dimensions of the components when the regression function is a composition of smooth, low-dimensional functions; see \citet{poggio2017why} for approximation-theoretic antecedents.  These analyses take the compositional structure of the target as a primitive assumption.  Our results are complementary in two ways.  First, Theorem \ref{thm:covering} quantifies the entropy of the compositional \emph{class} as a function of the correction radius \(\eps\), rather than fixing a structural decomposition of the target.  Second, Proposition \ref{prop:geometry}(iii) supplies a geometric mechanism---data concentrated near a low-curvature manifold---under which the target provably admits the compositional representation that such analyses assume.

\paragraph{Geometric stability and manifold models.}
\citet{bietti2021deep} show that invariance and geometric stability can improve sample complexity.  In nonparametric regression, manifold-adaptive methods achieve rates governed by the intrinsic dimension when the design is supported on a low-dimensional manifold \citep{bickel2007local}, and manifold structure underlies classical dimensionality-reduction and topology-recovery guarantees \citep{belkin2003laplacian,niyogi2008finding}.  Our contribution is complementary: instead of beginning with an invariant kernel, we analyze a compositional function class and show how curvature controls the size of the nonlinear correction required beyond a tangent linear representation.  This gives an explicit geometric route to smaller entropy.  Section \ref{sec:separation} complements these upper bounds with a class-level separation: on curved data, the affine compositional prior provably underperforms the matching curved model class---a statement about priors that is weaker than a minimax lower bound but is proved, not conjectured.

\paragraph{Neural networks and implicit priors.}
Deep networks impose compositional structure through architecture and optimization.  However, their effective prior is generally implicit and depends on initialization, parameterization, and training dynamics \citep{neyshabur2015path,bartlett2017spectrally,jacot2018neural}, and trained networks are observed to learn low-dimensional features through mechanisms that remain only partially understood \citep{radhakrishnan2024mechanisms}.  Our analysis should therefore be read as a theory of explicit compositional function priors, with neural networks providing motivation rather than being directly covered by the theorems.

\section{Conclusion}

We presented a function-prior framework for understanding when compositional structure reduces sample complexity.  For models \(f(x)=g(Ax+\Delta h(x))\), metric entropy decomposes into a finite-dimensional linear-map term, an intrinsic-dimensional outer-function term, and an ambient-dimensional correction term weighted by \(\eps\).  Geometry explains when \(\eps\) is small: locally low-curvature manifolds are well approximated by tangent linear maps, requiring only small nonlinear corrections.  Theorem \ref{thm:risk} composes these pieces into a finite-sample risk bound in which curvature enters only through \(\eps\), providing a quantitative account of when compositional priors behave as low-dimensional models.  Section \ref{sec:separation} adds the converse for a one-dimensional model problem: the correction budget is necessary, not merely sufficient---on curved data the affine prior faces an error floor that the matching curved model class does not, and in any dimension no affine feature map with a smaller budget serves every target on a curved chart.

\section*{Limitations}

The upper bounds are not claimed to be minimax optimal for the full compositional class, although Remark \ref{rem:lower} shows the ambient term reflects genuine capacity.  The excess-risk bound of Theorem \ref{thm:risk} is achieved by a theoretical sieve estimator rather than by the penalized estimator of Proposition \ref{prop:representer}, and the scale choice of Remark \ref{rem:r0choice} is analyzed only up to constants; data-driven selection of $r_0$ is not addressed.  The geometric construction requires $C^{s_h+1}$ regularity of the manifold beyond a bare curvature bound, and assumes the target factors through the nearest-point projection on a tube around the manifold.  Section \ref{sec:separation} proves a class-level floor and separation for a one-dimensional model problem and, in any dimension, a representation floor with the target chosen after the feature map; for a fixed scalar target the floor can fail when $k\ge2$ (the counterexample there), and a floor for fixed scalar targets with rotating level sets is not proved.  Minimax lower bounds over all estimators for the compositional class remain open; on curved data no such bound at the ambient rate can hold, since the circle-model sieve of Theorem \ref{thm:separation} attains the intrinsic rate.

\bibliographystyle{plainnat}
\bibliography{references}

\appendix

\section{Details for Theorem \ref{thm:covering}}
\label{app:covering}

Let \(h=Ax+\Delta h(x)\) and \(\widetilde h=\widetilde A x+\widetilde{\Delta h}(x)\).  By boundedness of \(\X\),
\[
\norm{(A-\widetilde A)x}_{L_n^2}
\le R\norm{A-\widetilde A}_F.
\]
If \(g\) is \(L_g\)-Lipschitz, then
\[
\norm{g\circ h-g\circ \widetilde h}_{L_n^2}
\le L_g\left(R\norm{A-\widetilde A}_F+\norm{\Delta h-\widetilde{\Delta h}}_{L_n^2}\right).
\]
For the outer function, after constructing a cover of the possible feature values in the bounded feature domain, choose \(\widetilde g\) satisfying
\[
\norm{g-\widetilde g}_{L_n^2(\widetilde h(x_1),\ldots,\widetilde h(x_n))}\le \delta_g.
\]
Thus
\[
\norm{g\circ h-\widetilde g\circ \widetilde h}_{L_n^2}
\le L_gR\norm{A-\widetilde A}_F+L_g\norm{\Delta h-\widetilde{\Delta h}}_{L_n^2}+\delta_g.
\]
Take the three terms to be at most \(\delta/3\).  The matrix cover contributes
\[
\log N\left(\frac{\delta}{3L_gR},\{A:\norm{A}_F\le B_A\},\norm{\cdot}_F\right)
\lesssim kd\log\left(\frac{C B_A L_g R}{\delta}\right).
\]
The Sobolev entropy bounds on bounded domains give
\[
\log N\left(\frac{\delta}{3L_g},\eps B_{W^{s_h,2}},L_n^2\right)
\lesssim \left(\frac{C L_g\eps}{\delta}\right)^{d/s_h},
\]
and
\[
\log N\left(\frac{\delta}{3},B_g B_{W^{s_g,2}},L_n^2\right)
\lesssim \left(\frac{C B_g}{\delta}\right)^{k/s_g}.
\]
Using \(L_g\lesssim B_g\) under the Sobolev embedding assumption gives the displayed bound in Theorem \ref{thm:covering}.  Constants may depend on domain regularity and smoothness parameters.

\paragraph{Bookkeeping.}
Four points used above deserve explicit statement.
\begin{enumerate}
\item[(i)] \emph{(Product structure of the cover.)}  The cover of the outer class is constructed separately for each element \((\widetilde A,\widetilde{\Delta h})\) of the covers of the matrix ball and of the correction class, so the total cover size is the product of the three cardinalities and the log-covering numbers add; the bound \eqref{eq:covering-bound} is exactly this sum.
\item[(ii)] \emph{(Bounded feature domain.)}  By the Sobolev embedding \(W^{s_h,2}\hookrightarrow L^\infty\), valid precisely because \(s_h>d/2\), every correction satisfies \(\sup_x\norm{\Delta h(x)}\le C\eps\), so all feature values lie in a ball of radius \(B_AR+C\eps\).  This is where the smoothness requirement of Assumption \ref{assump:smooth} enters the covering argument: it makes a single bounded feature domain available for the outer cover, uniformly over the class.
\item[(iii)] \emph{(Vector-valued correction.)}  \(\Delta h\) has \(k\) components, each in a scalar Sobolev ball, so its entropy carries a factor \(k\) relative to the scalar case; this factor is absorbed into the constants depending on \(k\).
\item[(iv)] \emph{(Uniformity over samples.)}  The empirical-\(L^2\) entropy bounds are applied uniformly over sample configurations.  This is justified by deriving them from sup-norm entropy bounds, which dominate every empirical \(L^2\) norm and which hold for the two Sobolev balls because \(s_h>d/2\) and \(s_g>k/2\).
\end{enumerate}

\begin{corollary}[Sup-norm covering]
\label{cor:supnorm}
Under Assumptions \ref{assump:domain}--\ref{assump:smooth}, the bound \eqref{eq:covering-bound} holds also for the uniform norm:
\[
\log N\bigl(\delta,\F_{B_A,B_g,\eps},\norm{\cdot}_\infty\bigr)
\lesssim
kd\log\!\left(\frac{C B_A B_g}{\delta}\right)
+
\left(\frac{C B_g\eps}{\delta}\right)^{d/s_h}
+
\left(\frac{C B_g}{\delta}\right)^{k/s_g}.
\]
\end{corollary}

\begin{proof}
Every inequality in the argument above is pointwise in \(x\): \(\norm{(A-\widetilde A)x}\le R\norm{A-\widetilde A}_F\) for all \(x\in\X\), and the composition and triangle inequalities hold pointwise.  The component classes admit uniform-norm covers of the same entropy orders: the two Sobolev balls because \(s_h>d/2\) and \(s_g>k/2\) (classical entropy of Sobolev balls in the uniform norm), and the matrix ball as before.  The product construction of item (i) is unchanged.
\end{proof}

\section{Details for Proposition \ref{prop:geometry}}
\label{app:geometry}

Throughout, work in the adapted coordinates of Assumption \ref{assump:geom}: \(\bar x=0\), \(T_{\bar x}M=\R^k\times\{0\}\), \(\varphi(u)=(u,\psi(u))\) for \(u\in B(0,r_0)\subset\R^k\), with \(\psi(0)=0\), \(D\psi(0)=0\), and \(\norm{D^m\psi}\le\kappa_m\le\bar\kappa\, r_0^{\,2-m}\) for \(2\le m\le s_h+1\).  All constants below may depend on \(d,k,s_h\) and the fixed cutoff.  The constant \(c_0=c_0(d,k,s_h)\) of Proposition \ref{prop:geometry} is fixed here: it is taken small enough that the inverse function theorem applies to the rescaled tube map in Step 1 (the only place its size is used quantitatively), and in any case \(c_0\le1/10\), as the proof of Corollary \ref{cor:atlas} requires.

\paragraph{Step 0: rescaling.}
Set \(x'=x/r_0\) and let \(M'=M/r_0\) be the rescaled manifold, the graph of \(\psi'(u')=\psi(r_0u')/r_0\) over \(B(0,1)\).  Then \(D^m\psi'(u')=r_0^{\,m-1}(D^m\psi)(r_0u')\), so
\[
    \norm{D^m\psi'}\le r_0^{\,m-1}\kappa_m\le\bar\kappa r_0\le c_0,
    \qquad 2\le m\le s_h+1,
\]
and \(\norm{D\psi'}\le\kappa_2r_0\le\bar\kappa r_0\) by the mean value theorem from \(D\psi'(0)=0\).  Thus in rescaled coordinates the manifold is \(\bar\kappa r_0\)-close to the flat \(\R^k\times\{0\}\) in \(C^{s_h+1}(B(0,1))\), the tube radius is \(\tau'=\tau/r_0\le1\), and the reach of \(M'\) is \(\rho/r_0\ge2\tau'\).  Functions transfer by \(w'(x')=w(r_0x')/r_0\), so that \(\sup\norm{D^mw}=r_0^{\,1-m}\sup\norm{D^mw'}\).  All bounds below are proved for the rescaled objects and transferred back at the end.

\paragraph{Step 1: frame and tube map.}
Let \(\nu'(u')\) be the orthonormal frame of \(N_{\varphi'(u')}M'\) obtained by Gram--Schmidt from a constant frame; it is a smooth algebraic function of \(D\psi'\), so \(\norm{D^m\nu'}\le C\bar\kappa r_0\) for \(1\le m\le s_h\) (every derivative carries at least one derivative of \(D\psi'\), and all factors are bounded since \(c_0\le1\)).  The tube map \(\Psi'(u',v')=\varphi'(u')+\nu'(u')v'\) satisfies \(\norm{D\Psi'-I}\le C\bar\kappa r_0\) and \(\norm{D^m\Psi'}\le C\bar\kappa r_0\) for \(2\le m\le s_h\).  For \(c_0\) small the inverse function theorem applies on the tube; by the Fa\`a di Bruno formula the derivatives of the inverse are polynomials in the derivatives of \(\Psi'\) and in \((D\Psi')^{-1}\), whence \(\norm{D\Psi'^{-1}}\le2\) and
\[
    \norm{D^m\Psi'^{-1}}\le C\bar\kappa r_0,\qquad 2\le m\le s_h:
\]
in rescaled coordinates the higher derivatives of the inverse are \emph{small}, not merely bounded.  Since \(\tau'\) is below the reach of \(M'\), the nearest-point projection satisfies \(\pi'(\Psi'(u',v'))=\varphi'(u')\), and the chart-coordinate function \(x'\mapsto u'(x')\) is \(C^{s_h}\) on the tube; one order of differentiability is lost relative to the \(C^{s_h+1}\) regularity of \(M\).

\paragraph{Step 2: the pre-correction and its derivatives.}
In the adapted coordinates \(A\varphi'(u')=u'\), so for \(x'=\Psi'(u',v')\) in the tube the pre-correction is exactly
\[
    w'(x')\;:=\;u'(\pi'(x'))-Ax'
    \;=\;u'-A\bigl(\varphi'(u')+\nu'(u')v'\bigr)
    \;=\;-\,A\nu'(u')v'.
\]
Since the rows of \(A\) span \(\R^k\times\{0\}\) and \(\nu'(0)\) is orthogonal to them,
\(\norm{A\nu'(u')}\le\norm{\nu'(u')-\nu'(0)}\le C\bar\kappa r_0\norm{u'}\), and \(\norm{D^m(A\nu')}\le C\bar\kappa r_0\) for \(1\le m\le s_h\).  Composing with \((u',v')=\Psi'^{-1}(x')\) and applying the chain and Leibniz rules, every term of \(D^m_{x'}w'\) contains at least one factor bounded by \(C\bar\kappa r_0\)---a value or derivative of \(A\nu'\), or a higher derivative of \(\Psi'^{-1}\)---while the remaining factors (\(\norm{v'}\le1\), first derivatives of \(\Psi'^{-1}\)) are bounded by absolute constants; additional small factors only help since \(c_0\le1\).  Hence
\begin{gather*}
    \sup\norm{D^mw'}\le C\bar\kappa r_0,\qquad 0\le m\le s_h,\\
    \text{and}\qquad
    \norm{w'}\le\norm{A\nu'}\norm{v'}\le C\bar\kappa r_0\,(\tau/r_0)=C\bar\kappa\tau .
\end{gather*}
Transferring back, \(\sup\norm{D^mw}\le C\bar\kappa r_0^{\,2-m}\) for \(0\le m\le s_h\), and \(\norm{w}_{L^\infty}\le C\bar\kappa r_0\tau\), the bound used in Remark \ref{rem:necessity}.

\paragraph{Step 3: extension and cutoff.}
Extend \(w'\) from the closed tube to \(\R^d\) by a Whitney extension preserving the \(C^{s_h}\) sup-norm bounds up to constants depending only on \((s_h,d)\)---in particular independent of \(r_0\)---and set \(\Delta h'(x'):=\chi(x')\,\widetilde w'(x')\), where \(\chi\) is a fixed smooth cutoff equal to \(1\) on \(B(0,7/4)\) and supported in \(B(0,2)\).  Every \(x'\) in the rescaled tube satisfies \(\norm{x'}\le(1+Cc_0)/2+\tau'<\tfrac74\), so the cutoff equals \(1\) there, giving property (i).  By the Leibniz rule \(\sup\norm{D^m\Delta h'}\le C\bar\kappa r_0\).  Back in the original coordinates, \(\Delta h(x)=r_0\Delta h'(x/r_0)\) is supported in \(B(0,2r_0)\) with \(\sup\norm{D^m\Delta h}\le C\bar\kappa r_0^{\,2-m}\), and integrating over the support volume,
\[
    \norm{D^m\Delta h}_{L^2(\R^d)}
    \le C\bar\kappa\, r_0^{\,2-m}\cdot r_0^{\,d/2},
    \qquad 0\le m\le s_h,
\]
which is \eqref{eq:curvature-bound-safe}; summing over \(m\le s_h\) yields the stated bound on \(\eps\).

\paragraph{Step 4: the chart property.}
On the manifold, \(h(\varphi(u))=u\), and \(\norm{\varphi(u)-\varphi(u')}=(1\pm C\bar\kappa r_0)\norm{u-u'}\) since \(\norm{D\psi}\le\kappa_2r_0\le\bar\kappa r_0\); this is the bi-Lipschitz claim in (i).

\paragraph{Step 5: factorization.}
Let \(G:=f_0\circ\varphi\) with \(\norm{G}_{C^{s_g}(B(0,r_0))}\le B_g\).  By the Whitney extension theorem there is \(EG\in C^{s_g}(\R^k)\) with \(EG=G\) on \(B(0,r_0)\) and \(\norm{EG}_{C^{s_g}(\R^k)}\le C(s_g,k)\,B_g\); Whitney's constants depend only on the smoothness order and the dimension, not on the domain.  Let \(\zeta\) be a fixed smooth cutoff equal to \(1\) on the feature domain (a ball of radius \(O(\sqrt kR)\), by Bookkeeping (ii) of Appendix \ref{app:covering}) and compactly supported; set \(g:=\zeta\cdot EG\).  Then \(g\) is compactly supported with
\[
    \norm{g}_{W^{s_g,2}(\R^k)}\le C(s_g,k,R)\,B_g,
\]
with constants independent of \(r_0\).  By contrast, an \(L^2\)-Sobolev (Stein) extension from \(B(0,r_0)\) would carry constants growing as \(r_0\to0\) (extending a constant function already forces \(C(r_0)\gtrsim r_0^{-k/2}\)); stating the chart hypothesis in the uniform norm avoids this, which matters for the scale analysis of Section \ref{sec:risk}.  Finally, for \(x\in T_{r_0,\tau}\), property (i) gives \(h(x)=u(\pi(x))\in B(0,r_0/2)\), where \(g=EG=G\), and hence
\[
    g(h(x))=G(u(\pi(x)))=f_0\bigl(\varphi(u(\pi(x)))\bigr)=f_0(\pi(x))=f(x).
\]
Constants throughout depend on the cutoffs and on the Whitney extension operators, and the dependence of all bounds on \(\bar\kappa\) is linear.

\section{Details for Theorem \ref{thm:risk}}
\label{app:risk}

The core of the proof is a generic bound for sieve regression over a finite sup-norm net; we state it as a standalone lemma because it is used again in Section \ref{sec:separation}.

\begin{lemma}[Sieve regression over a finite net]
\label{lem:sieve}
Let \((x_1,y_1),\ldots,(x_n,y_n)\) be i.i.d.\ with \(\E[y\mid x]=f^*(x)\) and \(|y|\le b\) almost surely.  Let \(G\) be a finite class of functions uniformly bounded by \(b\), containing an element \(f^\circ\) with \(\norm{f^\circ-f^*}_\infty\le\delta\), and let \(\hat f\) minimize the empirical squared loss over \(G\).  Then there is \(C=C(b)\) such that for every \(\eta\in(0,1)\), with probability at least \(1-\eta\),
\[
    \norm{\hat f-f^*}_{L^2(P)}^2
    \;\le\;C\Bigl(\delta^2+\frac{\log|G|+\log(2/\eta)}{n}\Bigr).
\]
\end{lemma}

\begin{proof}
\emph{Step 1: a Bernstein bound for excess losses.}
For \(f\in G\) define the excess loss
\(\ell_f(x,y):=(f(x)-y)^2-(f^*(x)-y)^2\).  Writing \((f-y)^2-(f^*-y)^2=(f-f^*)^2+2(f-f^*)(f^*-y)\) and using \(\E[y\mid x]=f^*(x)\),
\[
    P\ell_f=\norm{f-f^*}_{L^2(P)}^2\;\ge\;0 .
\]
Moreover \(|\ell_f|\le|f-f^*|\,|f+f^*-2y|\le8b^2\) and
\(\mathrm{Var}(\ell_f)\le\E\,\ell_f^2\le16b^2\,\norm{f-f^*}_{L^2(P)}^2=16b^2\,P\ell_f\):
the variance is linked to the excess risk.  Let \(t:=\log(2|G|/\eta)\).  By Bernstein's inequality and a union bound over the finite class \(G\), with probability at least \(1-\eta\), simultaneously for all \(f\in G\),
\[
    |P_n\ell_f-P\ell_f|
    \le\sqrt{\frac{32b^2\,P\ell_f\,t}{n}}+\frac{16b^2t}{3n}
    \le\tfrac14P\ell_f+\frac{Cb^2t}{n},
\]
using \(\sqrt{uv}\le\tfrac14u+v\) with \(u=P\ell_f\) and \(v=32b^2t/n\).  On this event, for every \(f\in G\),
\begin{equation}
    P\ell_f\le\tfrac43\,P_n\ell_f+\frac{Cb^2t}{n}
    \qquad\text{and}\qquad
    P_n\ell_f\le\tfrac54\,P\ell_f+\frac{Cb^2t}{n}.
    \label{eq:bernstein-pair}
\end{equation}

\emph{Step 2: bias.}
By hypothesis \(f^\circ\in G\) satisfies \(\norm{f^\circ-f^*}_\infty\le\delta\), whence \(P\ell_{f^\circ}=\norm{f^\circ-f^*}_{L^2(P)}^2\le\delta^2\).

\emph{Step 3: assembly.}
By definition of \(\hat f\), \(P_n\ell_{\hat f}\le P_n\ell_{f^\circ}\).  Combining with \eqref{eq:bernstein-pair},
\begin{align*}
    \norm{\hat f-f^*}_{L^2(P)}^2
    =P\ell_{\hat f}
    &\le\tfrac43P_n\ell_{\hat f}+\frac{Cb^2t}{n}
    \le\tfrac43P_n\ell_{f^\circ}+\frac{Cb^2t}{n}\\
    &\le\tfrac43\cdot\tfrac54\,P\ell_{f^\circ}+\frac{C'b^2t}{n}
    \le2\delta^2+\frac{C'b^2t}{n},
\end{align*}
with \(t=\log(2|G|/\eta)\) as above.
\end{proof}

\paragraph{Proof of Theorem \ref{thm:risk}.}
Apply Lemma \ref{lem:sieve} with \(G=G_\delta\), the internal \(\delta\)-net of Section \ref{sec:risk}: its members belong to \(\F^{\mathrm{atlas}}\) and hence share the uniform bound \(b\); it contains a point \(\delta\)-close to \(f^*\) in sup norm because \(f^*\in\F^{\mathrm{atlas}}\); and \(\log|G_\delta|\le\log N(\delta/2,\F^{\mathrm{atlas}},\norm{\cdot}_\infty)\).  This proves the first display of Theorem \ref{thm:risk}; the replacement of \(\delta\) by \(\delta/2\) inside the entropy affects only constants below.

\paragraph{Optimizing the net accuracy.}
By Corollaries \ref{cor:atlas} and \ref{cor:supnorm},
\begin{gather*}
    H(\delta):=\log N\bigl(\delta,\F^{\mathrm{atlas}},\norm{\cdot}_\infty\bigr)
    \;\le\;N\Bigl[kd\log\frac{C_0}{\delta}
    +\Bigl(\frac{C_1\eps}{\delta}\Bigr)^{p}
    +\Bigl(\frac{C_2}{\delta}\Bigr)^{q}\Bigr],\\
    p=\frac{d}{s_h}<2,\quad q=\frac{k}{s_g}<2,
\end{gather*}
with constants absorbing \(B_g,\sqrt k,R,b\).  Write \(H=H_1+H_2+H_3\) for the three terms and let \(\delta_i\) solve \(\delta_i^2=H_i(\delta_i)/n\):
\[
    \delta_1^2\asymp\frac{J\,kd\log n}{n},
    \qquad
    \delta_2^2\asymp\Bigl(\frac{J}{n}\Bigr)^{\frac{2s_h}{2s_h+d}}(C_1\eps)^{\frac{2d}{2s_h+d}},
    \qquad
    \delta_3^2\asymp\Bigl(\frac{J}{n}\Bigr)^{\frac{2s_g}{2s_g+k}}C_2^{\frac{2k}{2s_g+k}},
\]
where the first uses \(\log(C_0/\delta_1)\lesssim\log n\) for \(\delta_1\ge n^{-1}\), and the second and third follow from solving \(\delta^{2+p}=NC_1^p\eps^p/n\) and \(\delta^{2+q}=NC_2^q/n\) with \(2/(2+p)=2s_h/(2s_h+d)\) and \(2/(2+q)=2s_g/(2s_g+k)\).  Choose \(\delta:=\max(\delta_1,\delta_2,\delta_3)\).  Each \(H_i\) is nonincreasing, so \(H_i(\delta)/n\le H_i(\delta_i)/n=\delta_i^2\le\delta^2\), whence
\[
    \delta^2+\frac{H(\delta)}{n}\le4\delta^2\le4(\delta_1^2+\delta_2^2+\delta_3^2),
\]
which is \eqref{eq:risk-bound}.  Corollary \ref{cor:rates} follows by substituting \(\eps\le C\bar\kappa r_0^{\,d/2+2-s_h}\) (Proposition \ref{prop:geometry}(ii) with \(\max(r_0^{2-s_h},r_0^2)=r_0^{2-s_h}\) for \(r_0\le1\), \(s_h\ge2\)) and \(J\lesssim(D/r_0)^k\) (Corollary \ref{cor:atlas}).\hfill\(\square\)

\section{Proof of the shear lemma}
\label{app:separation}

\begin{proof}[Proof of Lemma \ref{lem:shear}]
Assume $a\ne0$ and rescale to $\norm{a}=1$ (level sets are
unchanged); let $w\perp a$ be the direction of the level lines.

\emph{Angular bookkeeping.}  Level sets and fibers are
\emph{lines}, so the relevant angle is taken modulo $\pi$, and the
correct transversality measure is $|\sin\beta(u)|$, where $\beta(u)$
is the angle between $w$ and the fiber direction $n(u)$: it vanishes
exactly when the level line is parallel to the fiber, regardless of
orientation.  For the circle, $\beta(u)=\beta(0)+\kappa u$ modulo
$\pi$, rotating at exact rate $\kappa$, with total rotation
$\kappa r_0\le c_0<\pi/2$ over the chart.  Since the zeros of
$\sin\beta(\cdot)$ are $\pi/\kappa$-separated in $u$, at most one
zero $u^*$ lies within angular reach of the chart; if none does, let
$u^*$ be the minimizer of $|\sin\beta|$ over the closed chart
interval (a boundary point).

\emph{Choice of a good strip.}  Removing the interval
$\{|u-u^*|<r_0/8\}$ from the chart leaves a set of length at least
$3r_0/4$ with at most two components, one of which contains an
interval $I'$ of length $r_0/4$.  On $I'$ the angular distance to
the nearest zero of $\sin\beta$ is at least $\kappa r_0/8$ and at
most $\pi/2$, so
\[
    |\sin\beta(u)|\;\ge\;\sin(\kappa r_0/8)\;\ge\;\tfrac{2}{\pi}\cdot
    \tfrac{\kappa r_0}{8}\;\ge\;\tfrac{\kappa r_0}{13},
\]
and $\sin\beta$ has constant sign on $I'$ (its only sign change in
reach is at $u^*$, which is excised).  Let
$T':=\{\varphi(u)+vn(u):u\in I',|v|\le\tau\}$; then
$P(T')\ge(1-Cc_0)\,|I'|/r_0\ge1/5$.

\emph{Monotone variation along level lines.}  Since
$\nabla(u\circ\pi)(x)=\frac{R}{\norm{x}}\,e_\theta(x)$ (valid for all
$x\ne0$) with $\frac{R}{\norm{x}}=1+O(\kappa\tau)$ on the relevant
region, along any level line
\[
    \frac{du}{ds}
    \;=\;\frac{R}{\norm{x}}\,\ip{e_\theta}{w}
    \;=\;\pm\frac{R}{\norm{x}}\,\sin\beta(u(x)) ,
\]
so at every point of the line with $u(x)\in I'$ we have
$|du/ds|\ge\kappa r_0/14=:m$, with constant sign.  Consequently $u$
is monotone in the line parameter on $\{x\in\text{line}:u(x)\in
I'\}$: if $u$ exited $I'$ and returned, it would have to reverse
direction inside $I'$, contradicting the fixed sign.  In particular,
between any two points of the line with $u\in I'$, the intermediate
points also have $u\in I'$, and $|u(x(s_1))-u(x(s_2))|\ge m|s_1-s_2|$.

\emph{Disconnected chords.}  Because $T'$ is an annular sector, a
level line nearly tangent to the inner boundary arc can meet $T'$ in
up to \emph{two} segments.  Let $s(t)$ be the \emph{total} length of
$\text{line}\cap T'$.  By the previous paragraph, $u$ is
$m$-bi-Lipschitz monotone in the line parameter across the whole
union (the gap between segments has $u\in I'$, since $u$ at both gap
endpoints lies in $I'$ and $u$ is monotone there).  For $S,S'$
i.i.d.\ uniform on the union,
\[
    \mathrm{Var}\bigl(u(S)\bigr)
    =\tfrac12\,\E\bigl(u(S)-u(S')\bigr)^2
    \ge\tfrac{m^2}{2}\,\E(S-S')^2
    = m^2\,\mathrm{Var}(S)
    \;\ge\; m^2\,\frac{s(t)^2}{12},
\]
where the last step uses that among unions of at most two intervals
of total length $s(t)$, the variance of the uniform measure is
minimized by the single contiguous interval.  Hence
$\mathrm{sd}'(f\mid t)\ge m\,s(t)/\sqrt{12}$.

\emph{Mass accounting.}  By Fubini in the rotated frame,
$\int s(t)\,dt=\mathrm{area}(T')=(1\pm Cc_0)\,|I'|\cdot2\tau
\ge\tfrac25\,r_0\tau$, while the $t$-interval over which lines meet
$T'$ has length at most
$\mathrm{diam}(T')\le|I'|+2\tau+Cc_0r_0\le\tfrac52 r_0$
(using $\tau\le r_0$).  Lines with $s(t)<\tau/16$ therefore
contribute at most $\tfrac{\tau}{16}\cdot\tfrac52 r_0=\tfrac{5}{32}
r_0\tau$ to the integral, so lines with $s(t)\ge\tau/16$ carry area
at least $\tfrac25 r_0\tau-\tfrac5{32}r_0\tau=\tfrac{39}{160}r_0\tau$,
which is at least a $\tfrac25$-fraction of
$\mathrm{area}(T')\le(1+Cc_0)r_0\tau/2$.  On those lines,
\[
    \mathrm{sd}'(f\mid t)
    \;\ge\;\frac{m\,\tau/16}{\sqrt{12}}
    \;\ge\;\frac{\kappa r_0\tau}{800}\;=:\;c\,\kappa r_0\tau .
\qedhere
\]
\end{proof}

\begin{proof}[Proof of Theorem \ref{thm:repfloor}]
Reduce to $P=\mathrm{Unif}(T)$ at the cost of the factor $c_P$.
Write points of $T$ as $x=(1+v/R)\,y$ with $y=\pi(x)\in U$ and
$v\in[-\tau,\tau]$; in these coordinates the uniform measure on $T$
has density proportional to $(1+v/R)^k$, which depends on $v$ only.

\emph{Step 0: the kernel direction and the adversarial target.}
Since $A$ has $k$ rows and $k+1$ columns, $\ker A\ne\{0\}$.  Fix a
unit $w\in\ker A$, let $q\in\{Rw,-Rw\}$ be the point nearer to $p$,
so that $d_q(p)\le\pi R/2$, and set $D:=\max(d_q(p),r_0)$ and
$f:=f_q$.  For $y\in U$, $d_q(y)\le d_q(p)+r_0/2$, hence
$\kappa d_q(y)\le\pi/2+c_0/2<\pi$: the function $d_q^2$ is smooth on
a neighborhood of $U$, with
$\nabla_M(d_q^2/2)(y)=d_q(y)\,\tau_y$, where $\tau_y$ is the unit
tangent at $y$ to the geodesic from $q$ through $y$, pointing away
from $q$ (and $\nabla_M d_q^2(q)=0$).  Thus $|\nabla_Mf|=d_q/D\le
(D+r_0/2)/D\le3/2$ on $U$, so $|f|\le3r_0/4$ there since $f(p)=0$,
and the derivatives of $d_q^2/2$ of order $m\ge2$ in geodesic
coordinates are bounded by $C(k,m)R^{2-m}=C(k,m)\kappa^{m-2}$;
dividing by $D\ge r_0$ and using $\kappa r_0\le1$ gives
$\norm{f}_{C^{s_g}(U)}\le C(k,s_g)(r_0+r_0^{1-s_g})=B_T$, so
$f\in\mathcal T$.

\emph{Step 1: near-constancy on lines parallel to $w$.}
Let $F=g(Ax+\Delta h(x))$ with $\mathrm{Lip}(g)\le L_g$.  On any line
$\ell$ parallel to $w$, $Ax$ is constant, so the feature ranges in an
interval of length at most $2\eps_\infty$ on $\ell\cap T$, and $F$
oscillates there by at most $\eta:=2L_g\eps_\infty$.

\emph{Step 2: disintegration along lines.}
Parametrize the lines parallel to $w$ by $t\in w^\perp\cong\R^k$,
$\ell_t:=t+\R w$.  For measurable $T'\subset T$, Fubini in the
orthonormal frame $(w,w^\perp)$ disintegrates $\mathrm{Unif}(T')$
into the normalized Lebesgue measures on $\ell_t\cap T'$, with mixing
law $\mu'$ on $t$ of density proportional to
$L(t):=\lambda(\ell_t\cap T')$.  Writing $\mathrm{sd}'(f\mid t)$ for
the standard deviation of $f$ under the conditional law on
$\ell_t\cap T'$, the proof of Lemma \ref{lem:disint}, which uses only
Step 1 on each cell, gives
\begin{equation}
    \E_{\mathrm{Unif}(T')}\bigl[(F-f)^2\bigr]
    \;\ge\;\int\bigl(\mathrm{sd}'(f\mid t)-\eta\bigr)_+^2\,d\mu'(t).
    \label{eq:repfloor-disint}
\end{equation}

\emph{Step 3: the target varies along the lines.}
For $x\ne0$ with $y=\pi(x)=Rx/\norm x$, $D\pi(x)=(R/\norm x)\,P_{T_yM}$,
so
\[
    \frac{d}{ds}f(x+sw)\Big|_{s=0}
    \;=\;\frac{R}{\norm x}\,\Bigl\langle\nabla_M\frac{d_q^2}{2D}(y),\,w\Bigr\rangle
    \;=\;\frac{R}{\norm x}\,\frac{d_q(y)}{D}\,\ip{\tau_y}{w}.
\]
Let $\theta:=\kappa d_q(y)$.  In the plane spanned by $y/R$ and
$\tau_y$, the unit vector toward $q$ is
$q/R=\cos\theta\,(y/R)-\sin\theta\,\tau_y$, and $w=\pm q/R$, so
$\ip{\tau_y}{w}=\mp\sin\theta$ and
\begin{equation}
    \frac{d}{ds}f(x+sw)
    \;=\;\mp\,\frac{R}{\norm x}\,\frac{d_q(y)}{D}\,\sin\bigl(\kappa d_q(y)\bigr),
    \label{eq:repfloor-derivative}
\end{equation}
which has constant sign wherever $0\le\kappa d_q(\pi(x))\le\pi$.  Let
$T':=\{x\in T:\ d_q(\pi(x))\ge r_0/8\}$.  On $T'$ the magnitude of
\eqref{eq:repfloor-derivative} is at least $m:=\kappa r_0/200$.
Indeed $R/\norm x\ge(1+\kappa\tau)^{-1}\ge2/3$, and
$\sin\theta\ge\tfrac12\min(\theta,1)$ on $[0,\pi/2+c_0/2]$.  If
$d_q(p)\le r_0$ then $D=r_0$ and $d_q(y)\in[r_0/8,3r_0/2]$ on $T'$,
so $\theta\le3c_0/2\le1$ and
$(d_q/D)\sin\theta\ge\tfrac18\cdot\tfrac12\cdot\tfrac{\kappa r_0}8=\kappa r_0/128$.
If $d_q(p)>r_0$ then $D=d_q(p)$ and $d_q(y)\ge D-r_0/2\ge D/2$, so
$(d_q/D)\sin\theta\ge\tfrac12\cdot\tfrac12\min(\kappa D/2,1)\ge\tfrac14\cdot\tfrac{\kappa r_0}2=\kappa r_0/8$.
In both cases the magnitude is at least $\tfrac23\cdot\kappa r_0/128\ge m$.
The sub-tube carries mass
$P(T')=\mathrm{vol}_M(U\setminus B_M(q,r_0/8))/\mathrm{vol}_M(U)$,
because the density $(1+v/R)^k$ integrates out identically over $v$;
geodesic balls of radius $r\le r_0$ in $S^k(R)$ have volume between
$(1+c_0^2)^{-k}\omega_kr^k$ and $\omega_kr^k$, so
$\mathrm{vol}_M(B_M(q,r_0/8))/\mathrm{vol}_M(U)\le(1+c_0^2)^k4^{-k}\le\tfrac13$
and $P(T')\ge\tfrac23$.

\emph{Step 4: variance along a line.}
Fix $t$ with $L(t)>0$ and let $G_t:=\{s:\ t+sw\in T'\}$.  The
function $s\mapsto f(t+sw)$ is monotone on $\{s:\ t+sw\in\mathrm{conv}(T)\}$:
$T$ lies in the half-space $\{\ip{x}{p/R}\ge(R-\tau)\cos(\kappa r_0/2)\}$,
which excludes the origin, so the segment between two points of $T$
projects radially onto the minor great-circle arc between their
projections, which stays in the geodesically convex set $U$; hence
$\kappa d_q(\pi(x))<\pi$ along the segment and
\eqref{eq:repfloor-derivative} keeps its sign.  Consequently, for
$s_1<s_2$ in $G_t$,
$|f(t+s_2w)-f(t+s_1w)|\ge m\,\lambda(G_t\cap[s_1,s_2])$.
Let $S,S'$ be independent and uniform on $G_t$ and set
$G(s):=\lambda(G_t\cap(-\infty,s])$, so that $G(S)$ is uniform on
$[0,L(t)]$.  Then
\[
    \mathrm{Var}\bigl(f(t+Sw)\bigr)
    =\tfrac12\,\E\bigl(f(t+Sw)-f(t+S'w)\bigr)^2
    \ge\tfrac{m^2}2\,\E\bigl(G(S)-G(S')\bigr)^2
    =\frac{m^2L(t)^2}{12},
\]
i.e.\ $\mathrm{sd}'(f\mid t)\ge mL(t)/\sqrt{12}$.  No count of the
segments of $G_t$ is needed.

\emph{Step 5: mass accounting.}
By Fubini, $\int L(t)\,dt=\mathrm{vol}(T')\ge\tfrac23\mathrm{vol}(T)$,
and $\mathrm{vol}(T)=\mathrm{vol}_M(U)\int_{-\tau}^{\tau}(1+v/R)^kdv\ge c_kr_0^k\tau$
with $c_k$ depending only on $k$.  The lines meeting $T'$ have $t$ in
the orthogonal projection of $T$ onto $w^\perp$, which lies in a
$k$-dimensional ball of radius $\mathrm{diam}(T)/2\le2r_0$ and so has
$k$-volume at most $C_kr_0^k:=\omega_k(2r_0)^k$.  Let
$\lambda_0:=c_k\tau/(3C_k)$.  Lines with $L(t)<\lambda_0$ contribute
at most $\lambda_0C_kr_0^k=c_kr_0^k\tau/3\le\tfrac12\mathrm{vol}(T')$
to $\int L\,dt$, so the lines with $L(t)\ge\lambda_0$ carry
$\mu'$-mass at least $\tfrac12$, and on them
$\mathrm{sd}'(f\mid t)\ge m\lambda_0/\sqrt{12}=:c_k'\,\kappa r_0\tau$.

\emph{Step 6: assembly.}
Since the integrand is nonnegative,
\[
    \E_P\bigl[(F-f)^2\bigr]
    \;\ge\;P(T')\,\E_{\mathrm{Unif}(T')}\bigl[(F-f)^2\bigr]
    \;\ge\;\tfrac23\cdot\tfrac12\,
    \bigl(c_k'\,\kappa r_0\tau-2L_g\eps_\infty\bigr)_+^2
\]
by \eqref{eq:repfloor-disint} and Steps 4--5.  The function $F$ was
an arbitrary member of the affine class with the given $A$ and
$\Delta h$, and $f=f_q\in\mathcal T$ depends only on $A$; this proves
\eqref{eq:repfloor} with $c=c_k'^{\,2}/3$ and $C=2/c_k'$.  The
Sobolev version follows exactly as in Theorem \ref{thm:floor}.
\end{proof}

\begin{remark}[Higher codimension]
\label{rem:repfloor-codim}
For $M=S^k(R)\subset\R^{k+1}\times\{0\}\subset\R^d$ with $d>k+1$,
the tube is $\{y+v:\ y\in U,\ v\in N_yM,\ \norm v\le\tau\}$ with
$N_yM=\R y\oplus(\{0\}\times\R^{d-k-1})$, and the nearest-point
projection depends only on the first $k+1$ coordinates.  Since
$\dim\ker A\ge d-k$, the kernel contains a unit vector
$w\in\R^{k+1}\times\{0\}$; lines parallel to $w$ lie in the affine
slices $\{x''=\mathrm{const}\}$, each of which meets the tube in a
copy of the codimension-one problem with tube radius
$\sqrt{\tau^2-\norm{x''}^2}\ge\tau/2$ when $\norm{x''}\le\tfrac{\sqrt3}2\tau$,
and those slices carry a fraction of the mass depending only on
$d-k$.  Running the proof slice by slice gives Theorem
\ref{thm:repfloor} in any codimension, with constants depending on
$d$ and $k$.  We omit the details.
\end{remark}

\section{Notes on the compositional penalty}
\label{app:penalty}

The componentwise penalty \eqref{eq:additive-prior} is Hilbertian in \(A\), \(\Delta h\), and \(g\), but the induced functional on \(f\) need not be quadratic.  Even in a simple scalar linear submodel, scaling can be redistributed across layers.  For example, if \(f_c(x)=cx\) and one writes \(f_c(x)=g(ax)\) with \(g(z)=bz\), then \(ab=c\) and the two-layer cost \(a^2+b^2\) is minimized at scale \(2|c|\), not at scale \(c^2\).  Thus the natural induced penalty from optimizing over decompositions can fail the homogeneity expected of a quadratic norm.  This is the sense in which compositional function priors leave the ordinary single-RKHS regime.

\paragraph{Quadratic versus compositional priors.}
Kernel ridge regression is the special case in which \(\Phi(f)=\norm{f}_{\Hcal}^2\) for an RKHS \(\Hcal\).  In that case, the representer theorem reduces an infinite-dimensional problem to a finite-dimensional one, and spectral methods give sharp generalization theory in many settings \citep{scholkopf2002learning,bach2024learning}.  We do not claim that every quadratic-looking functional is equivalent to an RKHS norm without additional assumptions such as completeness and continuity of evaluation.  Rather, RKHS regularization is the canonical quadratic function-space prior.  The compositional penalty \eqref{eq:additive-prior} is different: even with Hilbertian component norms, optimization over decompositions and the nonlinear map \((g,h)\mapsto g\circ h\) produces a non-quadratic prior on \(f\).

\begin{proposition}[Componentwise representer form]
\label{prop:representer}
Consider the regularized empirical problem
\begin{equation}
\min_{A,\Delta h,g}
    \sum_{i=1}^n \bigl(g(Ax_i+\Delta h(x_i))-y_i\bigr)^2
    +\lambda_A\norm{A}_F^2+\lambda_h\norm{\Delta h}_{\Hcal_h}^2
    +\lambda_g\norm{g}_{\Hcal_g}^2 .
    \label{eq:regularized-problem}
\end{equation}
Assume \(\Hcal_h\) is a vector-valued RKHS with scalar kernel \(K_h\) applied componentwise, and \(\Hcal_g\) is an RKHS with kernel \(K_g\).  Any minimizer, if it exists, has a representation
\begin{align}
    \Delta h(x) &= \sum_{i=1}^n \alpha_i K_h(x_i,x), \qquad \alpha_i\in\R^k,\label{eq:repr-h}\\
    g(z) &= \sum_{j=1}^n \beta_j K_g(z_j,z), \qquad z_j=Ax_j+\Delta h(x_j),\label{eq:repr-g}
\end{align}
with \(A\in\R^{k\times d}\).  Thus the optimization is finite-dimensional, with \(kd+nk+n\) scalar parameters, but it remains nonconvex because the variables are coupled through \(z_j\).
\end{proposition}

\begin{proof}
For fixed \(A\) and \(\Delta h\), the objective depends on \(g\) only through evaluations \(g(z_1),\ldots,g(z_n)\) plus the squared RKHS norm.  The standard representer theorem gives \eqref{eq:repr-g}.  For fixed \(A\) and \(g\), the objective depends on \(\Delta h\) only through \(\Delta h(x_1),\ldots,\Delta h(x_n)\) plus the squared vector-valued RKHS norm.  The generalized representer theorem gives \eqref{eq:repr-h}.  The matrix \(A\) is already finite-dimensional.  The parameter count is therefore \(kd+nk+n\).  The proof establishes representation, not convexity or uniqueness.
\end{proof}

\begin{remark}[Existence of a minimizer]
\label{rem:existence}
Proposition \ref{prop:representer} is conditional on existence.  A minimizer of \eqref{eq:regularized-problem} exists whenever the kernels $K_h$ and $K_g$ are continuous, by the direct method.  \emph{Coercivity:} the three penalties bound $\norm{A}_F$, $\norm{\Delta h}_{\Hcal_h}$, and $\norm{g}_{\Hcal_g}$ along any minimizing sequence, which therefore has a subsequence converging weakly in $\R^{k\times d}\times\Hcal_h\times\Hcal_g$, bounded sets of a Hilbert space being weakly sequentially compact.  \emph{Lower semicontinuity:} the squared norms are weakly lower semicontinuous.  For the data term, if $(A_m,\Delta h_m,g_m)\rightharpoonup(A,\Delta h,g)$, then $z^m_j:=A_mx_j+\Delta h_m(x_j)\to z_j:=Ax_j+\Delta h(x_j)$, because evaluation at $x_j$ is a bounded functional on $\Hcal_h$, and
\[
    g_m(z^m_j)-g(z_j)
    \;=\;\bigl\langle g_m,\,K_g(z^m_j,\cdot)-K_g(z_j,\cdot)\bigr\rangle_{\Hcal_g}
    +\bigl\langle g_m-g,\,K_g(z_j,\cdot)\bigr\rangle_{\Hcal_g}
    \;\longrightarrow\;0,
\]
the first term because $\sup_m\norm{g_m}_{\Hcal_g}<\infty$ and $z\mapsto K_g(z,\cdot)$ is continuous into $\Hcal_g$ when $K_g$ is continuous (its squared increment is $K_g(z,z)-2K_g(z,z')+K_g(z',z')$), the second by weak convergence.  The objective is therefore weakly lower semicontinuous and coercive, and attains its infimum.  The minimizer need not be unique, and the finite-dimensional problem in $(A,\alpha,\beta)$ remains nonconvex; existence says that the representer form describes an actual optimum rather than an infimum that is not attained.  For the Sobolev classes of Section \ref{sec:risk} the question does not arise, since the sieve estimator minimizes over a finite set.
\end{remark}

\end{document}